%% file: main.tex
\documentclass[10pt,twocolumn,letterpaper]{article}

\usepackage{cvpr}              

\input{preamble}

\definecolor{cvprblue}{rgb}{0.21,0.49,0.74}
\usepackage[pagebackref,breaklinks,colorlinks,linkcolor=red, citecolor=cvprblue, urlcolor=magenta]{hyperref}

\usepackage{amsmath}
\usepackage{amsfonts} 
\usepackage{algorithmic}
\usepackage{array}
\usepackage{textcomp}
\usepackage{stfloats}
\usepackage{bm}
\usepackage{graphicx}
\usepackage{booktabs}
\usepackage{subcaption}
\usepackage{colortbl}
\usepackage{threeparttable}
\usepackage{makecell}
\usepackage{cancel}
\usepackage{bbding}
\usepackage{multirow}
\usepackage{tabularx}
\usepackage{footnote}
\usepackage{hyperref}
\usepackage{pifont}  
\usepackage{soul}

\definecolor{lightred}{RGB}{255, 230, 230}
\definecolor{lightgreen}{RGB}{230, 255, 230}
\definecolor{lightyellow}{RGB}{255, 255, 204}
\definecolor{lightorange}{RGB}{255, 229, 204}
\definecolor{lightblue}{RGB}{204, 229, 255}
\definecolor{limegreen}{RGB}{76, 175, 80} 
\definecolor{verylightgray}{gray}{0.95} 
\newcolumntype{L}[1]{>{\raggedright\arraybackslash}p{#1}}
\newcolumntype{C}[1]{>{\centering\arraybackslash}p{#1}}

\def\confName{CVPR}
\def\confYear{2025}

\title{EgoTextVQA: Towards Egocentric Scene-Text Aware Video Question Answering}

\author{
    Sheng Zhou\textsuperscript{1, 2}$^{*}$ \quad
    Junbin Xiao\textsuperscript{2}$^{\dagger}$ \quad
    Qingyun Li\textsuperscript{2} \quad
    Yicong Li\textsuperscript{2} \quad 
    Xun Yang\textsuperscript{3} \quad
    Dan Guo\textsuperscript{1} \quad \\
    Meng Wang\textsuperscript{1} \quad
    Tat-Seng Chua\textsuperscript{2} \quad
    Angela Yao\textsuperscript{2} \\
    \small \textsuperscript{1}Hefei University of Technology \quad
    \textsuperscript{2}National University of Singapore \quad
    \textsuperscript{3}University of Science and Technology of China \\
    \tt\small hzgn97@gmail.com, \tt\small \{junbin, ayao\}@comp.nus.edu.sg
}

\begin{document}

\twocolumn[{%
\renewcommand\twocolumn[1][]{#1}%
\maketitle
\begin{center}
    \centering
    \captionsetup{type=figure}
    \includegraphics[width=1.0\textwidth]{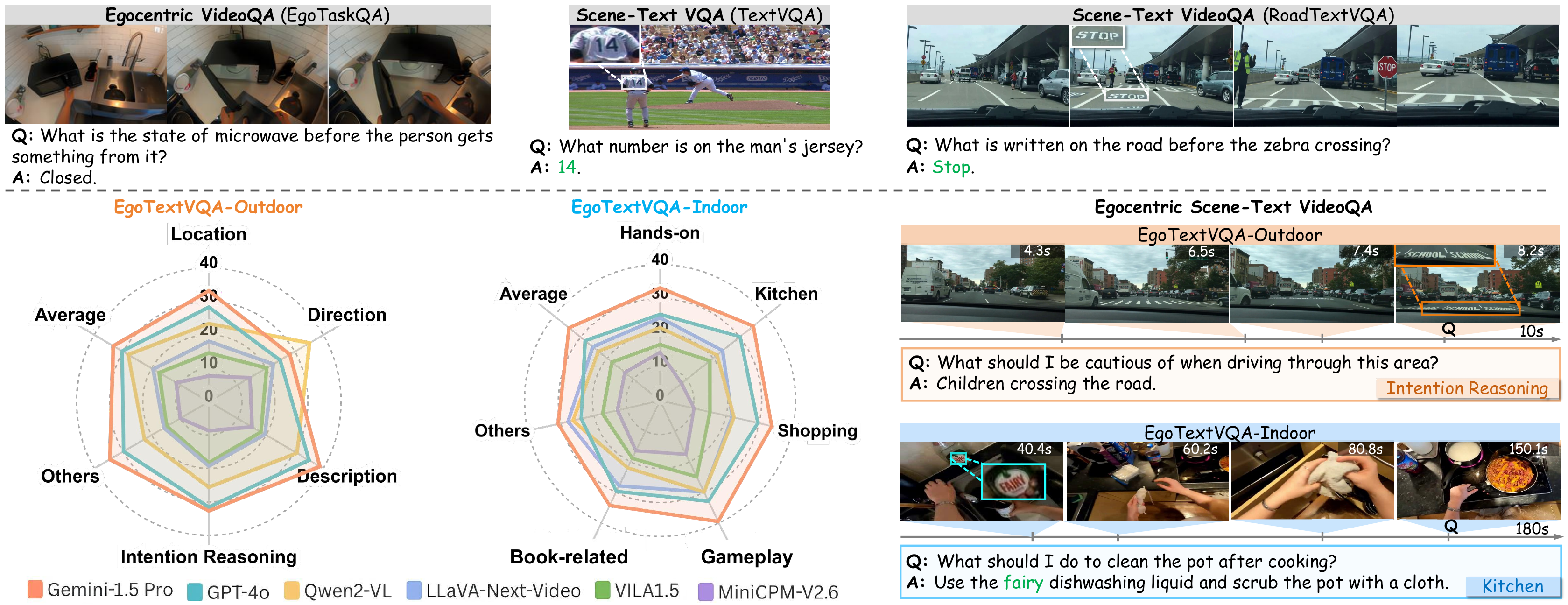}
    \captionof{figure}{
    Our \dataset~aims for QA assistance involving scene text from an ego-perspective mainly in outdoor driving (\datasetout) and indoor house-keeping (\datasetin), with the questions reflecting the real user needs yet without the visual focus on scene text. Benchmarking results show that all models struggle on \dataset, highlighting continued efforts for improvements. }
    \label{fig:intro}
\end{center}
}]

\maketitle
\renewcommand{\thefootnote}{}
\footnotetext{${}^*$ Work done when the first author was a visiting student at NUS.}
\footnotetext{$\dagger$ Corresponding author.}

\input{sec/0_abstract} 
\vspace{-0.4cm}
\input{sec/1_intro}

\input{sec/2_relatedwork}

\input{sec/3_benchmark}

\input{sec/4_experiment}
\vspace{-0.3cm}
\input{sec/5_conclusion}

\input{sec/6_acknowledgements}

{
    \small
    \bibliographystyle{ieeenat_fullname}
    \bibliography{main}
}

\input{sec/X_suppl}

\end{document}

%% file: preamble.tex
\newcommand{\cmark}{\textcolor{green}{\ding{51}}} 
\newcommand{\xmark}{\textcolor{red}{\ding{55}}} 
\newcommand{\dataset}{{EgoTextVQA}}
\newcommand{\datasetout}{{EgoTextVQA-Outdoor}}
\newcommand{\datasetin}{{EgoTextVQA-Indoor}}


%% file: sec/0_abstract.tex
\begin{abstract} 
We introduce EgoTextVQA, a novel and rigorously constructed benchmark for egocentric QA assistance involving scene text. EgoTextVQA contains 1.5$K$ ego-view videos and 7$K$ scene-text aware questions that reflect real user needs in outdoor driving and indoor house-keeping activities. The questions are designed to elicit identification and reasoning on scene text in an egocentric and dynamic environment. With EgoTextVQA, we comprehensively evaluate 10 prominent multimodal large language models. Currently, all models struggle, and the best results (Gemini 1.5 Pro) are around 33\% accuracy, highlighting the severe deficiency of these techniques in egocentric QA assistance. Our further investigations suggest that precise temporal grounding and multi-frame reasoning, along with high resolution and auxiliary scene-text inputs, are key for better performance. With thorough analyses and heuristic suggestions, we hope EgoTextVQA can serve as a solid testbed for research in egocentric scene-text QA assistance. Our dataset is released at: \href{https://github.com/zhousheng97/EgoTextVQA}  {https://github.com/zhousheng97/EgoTextVQA}.
\end{abstract}

%% file: sec/1_intro.tex
\section{Introduction}
\label{sec:intro}
The advances in wearable cameras and egocentric
vision research~\cite{fan2019egovqa, grauman2022ego4d, lin2022egocentric, pramanick2023egovlpv2} have led to a surge of interest in developing assistance applications. A promising direction is egocentric video question answering (VideoQA) \cite{xiao2024videoqa, li2023transformer, yang2024robust, guo2021context, di2024grounded, zhangmulti}, where AI agents offer live help by answering user questions. Despite the potential, related dataset research~\cite{ye2024mm, fan2019egovqa, jia2022egotaskqa, mangalam2023egoschema, cheng2024videgothink} has been designed to assess VideoQA models in their visual understanding capabilities. The questions are not curated to reflect real user needs in assistive applications (see EgoTaskQA in~Figure~\ref{fig:intro}). 
QAEgo4D \cite{Baermann_2022_CVPR} and AssistQ \cite{wong2022assistq} are closer in aim, but these datasets are designed to assist with episodic memory and ``how to'' demonstration; they emphasize pure visual understanding while neglecting scene text.  

Research in assistive techniques \cite{gurari2018vizwiz, zhou2023exploring, singh2019towards, li2022invariant} have shown that scene text, including signs, labels, and other text elements that are ubiquitous in our daily lives, is frequently involved in user questions when seeking assistance. Existing efforts~\cite{singh2019towards, zhou2024scene, biten2019scene} for QA on scene-text primarily aim at Optical Character Recognition (OCR) from ideal imaging conditions.  
They assume that people take good pictures and ask questions directly pinpointing the scene-text regions (see TextVQA~\cite{singh2019towards} in~Figure~\ref{fig:intro}). Despite the progress (\eg, leading results have accuracies of 85\% on ST-VQA \cite{biten2019scene,wang2024qwen2}), such a simplified setting has limited application value. Consider, for example, those with a visual impairment~\cite{gurari2018vizwiz}, they will likely struggle to take clean and in-focus images; or can they point to scene-text regions. More recent works \cite{zhao2022towards,tom2023reading} begin to study text-based VideoQA which allows QA on multiple images to reduce the proportion of unanswerable cases. Nonetheless, they still assume people know well the locations of scene text (see RoadTextVQA~\cite{tom2023reading} in~Figure~\ref{fig:intro}). Corresponding VQA models, while achieving good results on such benchmarks, cannot reason over user intentions other than an extraction of the OCR results \cite{jahagirdar2023understanding}.

In light of this, we introduce \dataset, a novel and rigorously constructed \underline{Ego}centric Scene-\underline{Text} Aware \underline{V}ideo \underline{Q}uestion \underline{A}nswering benchmark. \dataset ~is designed to advance research on egocentric QA assistance in real-life scenarios. We collect 1.5K ego-view videos and 7K scene-text aware questions from outdoor driving (\datasetout) and indoor house-keeping (\datasetin) activities. By emphasizing real user needs, we allow questions and answers that do not explicitly pinpoint the scene text. However, comprehending scene text is crucial to answering the questions (see examples in Figure \ref{fig:intro}). Moreover, to simulate real-time and streaming video QA, we set a timestamp for each question and allow the models to access \emph{only} visual contents before the question's timestamp to make related responses. Detailed visual scenarios and question types are also provided for better analysis of models.

For its realistic setting, \dataset~retains all the challenges in existing scene text VQA datasets and incorporates new major difficulties: First and foremost, it is crucial to \emph{reason about the user's intentions} with respect to the visual scene and scene text to both understand and to answer the questions. 
Second, the model may need to \emph{reason across multiple frames} to either infer the user's behavior or to locate the required scene text for answering. 
Third, the model may need to \emph{infer the user's current state} when posing the questions and provide actionable answers for meaningful assistance. Finally, the egocentric dynamic situation poses an additional challenge to \emph{scene text recognition} as opposed to the well-captured images and focused scene text.

With \dataset, we comprehensively benchmark 10 prominent models that perform well on existing scene-text VQA datasets, and find they all struggle, especially the open-source ones like MiniCPM-V 2.6 \cite{yao2024minicpm}, ShareGPT4Video \cite{chen2024sharegpt4video} and CogVLM2-Video \cite{hong2024cogvlm2}. The best-performing model is Gemini 1.5 Pro~\cite{reid2024gemini}, yet its accuracy is still lower at around 33\%. Comprehensive analyses and additional heuristic explorations reveal many insights on model behaviors and possible directions for improvements: 
\begin{enumerate}
    \item Temporally localizing the key frames and jointly reasoning over multiple key frames (\vs~a single key frame) are crucial for improvements, especially for long video QA scenarios on \datasetin.
    \item High-resolution image and scene text input can significantly boost models' performances. Yet, one needs to mind the compute efficiency. Also, models that inherently take fixed higher-resolution inputs are not necessarily better than those using lower-resolution inputs.
    \item Scene text from additional OCR techniques, despite being auxiliary, are extremely helpful for all models. 
\end{enumerate}

Additionally, we find that even humans cannot perform well on this task (43\% on \datasetout~and 27\% on \datasetin), largely
due to the difficulty of scene text recognition and the limited knowledge of humans, highlighting the significance of this research.
To summarize our contributions: 1) We propose to study scene-text aware VideoQA towards egocentric assistance and construct the \dataset ~dataset containing both \datasetin ~and \datasetout. 2) We benchmark 10 contemporary powerful models covering both open-source and closed-source ones and comprehensively analyze their limitations. 3) We explore a series of heuristic methods for improvements and share insightful findings for future work.

%% file: sec/2_relatedwork.tex
\section{Related Work}
\subsection{Scene-Text VQA Benchmark}
In the scene-text VQA field, various image- and video-based datasets~\cite{singh2019towards, wang2020general, biten2019scene, zhao2022towards, tom2023reading, mishra2019ocr, liu2024ocrbenchhiddenmysteryocr} have been proposed.  
The datasets like TextVQA~\cite{singh2019towards}, ST-VQA~\cite{biten2019scene}, and ESTVQA~\cite{wang2020general} provide high-quality images and questions that clearly point to scene text, but such simple settings limit practical applications. 
To expand the application scope, the datasets like M4viteVQA~\cite{zhao2022towards} and RoadTextVQA~\cite{tom2023reading} offer text-rich videos, but they design simple questions which refer to well-focused scene text. Such simplified settings make QA less practical and they challenge no more than an identification of the scene text. In addition, RoadTextVQA focuses merely on road driving.
In this work, we introduce \dataset~to advance research in real-life QA assistance from an egocentric perspective. Our videos cover diverse daily scenarios, with the questions reflecting real user needs yet without the visual attention on scene text.

\subsection{MLLMs for Scene-Text VQA}
Recent advancements in Multimodal Large Language Models (MLLMs)~\cite{zhang2024beyond, wang2023cogvlm, zhang2024llava} have shown significant potential in addressing scene-text VQA tasks~\cite{singh2019towards, biten2019scene, liu2024ocrbenchhiddenmysteryocr}.  Current MLLMs~\cite{luo2024feast, zhang2024beyond, li2025flexattention} achieve substantial performance gains in image scene-text QA by enhancing their ability to process higher-resolution images, leading to significant improvements and strong performance. However, these models are limited to image-level input, leaving their video comprehension capabilities largely unexplored. To address this limitation, recent models~\cite{yao2024minicpm, zhang2024llavanextvideo, wang2024qwen2, lin2024vila, chen2024far, yang2022video,yang2021deconfounded,hong2024cogvlm2,chen2024sharegpt4video} are developed to support video-level or multi-frame inputs while retain their performance on image-based scene-text QA datasets. Nonetheless, these models have been evaluated separately on image-based OCR-rich benchmarks~\cite{singh2019towards, biten2019scene, liu2024ocrbenchhiddenmysteryocr, mishra2019ocr, mathew2021docvqa, mathew2022infographicvqa} and video understanding benchmarks~\cite{mangalam2023egoschema, xiao2021next, fu2024video, li2024mvbench,shang2019annotating}, lacking suitable assessment for scene-text comprehension in more realistic and dynamic environments. In this work, we will comprehensively analyze the model behaviors in egocentric QA assistance in real-life dynamic environments.

%% file: sec/3_benchmark.tex
\section{\dataset~Dataset}

\begin{figure*}
\centering
\includegraphics[width=\textwidth]{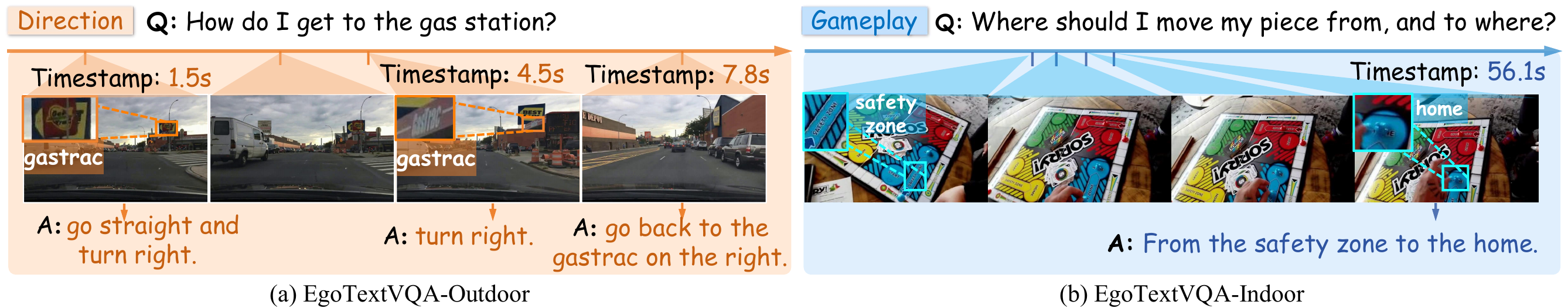}
\caption{Examples of \dataset. Scene text plays pivotal role in understanding and answering the questions which reflect real user needs. Yet, the videos are without the visual focus on scene text.} 
\label{fig:exp}
\vspace{-0.5cm}
\end{figure*}

\subsection{Dataset Creation} 
We leverage existing text-rich video sequences and the multimodal understanding and generation capabilities of MLLMs to create our dataset. The generated QA candidates are carefully reviewed and refined by human annotators as necessary to ensure the dataset's high diversity and quality.

\begin{table}
\centering
\caption{Data distribution and related examples of \dataset. The {\color{limegreen}green} words denote scene texts that appear in the video. }
\vspace{-0.2cm}
\label{tab:data_dis}
\setlength{\tabcolsep}{.8em}
\fontsize{6.4}{8}\selectfont
\begin{tabular}{>{\raggedright\arraybackslash}p{0.95cm} | >{\raggedright\arraybackslash}p{0.75cm} | >{\raggedright\arraybackslash}p{4.8cm}}  
\Xhline{1pt}
\textbf{Category} & \makecell[l]{\textbf{\#Q / \#V}} & \makecell[l]{\textbf{QA Examples}}   \\ 
\Xhline{1pt}
\rowcolor{lightorange}\multicolumn{3}{c}{{\datasetout}} \\
\hline
\textbf{Location} & \makecell[l]{1,505/553} & \makecell[l]{\emph{Q: Where should I go if I need a dentist?} \\ A: {\color{limegreen}Vijaya Dental Hospital}.} \\ 
\hline
{\textbf{Description}} & 1,387/601 & \makecell[l]{\emph{Q: What is the license plate of the car in front}  \emph{of me?} \\
A: {\color{limegreen}Ts07ge5554}.} \\ 
\hline
\textbf{Direction} & 952/468 & \makecell[l]{\emph{Q: Which direction do I go to get to {\color{limegreen}Vapor Inn}?} \\
A: Continue straight and turn right at the next road.} \\ 
\hline
\makecell[l]{\textbf{Intention} \\ \textbf{Reasoning}} & 813/443& \makecell[l]{\emph{Q: What should I be cautious of in this area?} \\ A: Watch for pedestrians crossing at the crosswalk.}  \\
\hline
\textbf{Others} & 191/152& \makecell[l]{\emph{Q: What is the main theme of the event advertised on } \\ \emph{the bus and when and where is it happening?} \\ A: The main theme is ``{\color{limegreen}Race Against HIV}", and it is  \\ happening on  {\color{limegreen}Dec 1, 2018}, at {\color{limegreen}People's Plaza}.} \\ 
\hline
\rowcolor{lightblue}\multicolumn{3}{c}{{\datasetin}} \\
\hline
\textbf{Hands-on} & 698/284 &  \makecell[l]{\emph{Q: How should I properly handle the {\color{limegreen}AVT} equipment?} \\ A: Maintain a firm grip and use proper posture. }   \\ 
\cline{1-3}
\textbf{Shopping} & 358/78 &  \makecell[l]{\emph{Q: Where can I find the ``{\color{limegreen}Milano}" cookies?}  \\ A: Top shelf on the right. }   \\ 
\cline{1-3}
\textbf{Kitchen} & 335/149 & \makecell[l]{\emph{Q: What should I use to clean up countertop spills? } \\ A: {\color{limegreen}Ecover} dish soap }   \\ 
\cline{1-3}
\textbf{Book-related} & 323/103 & \makecell[l]{\emph{Q: Where can I find information on learning the guitar? } \\ A: {\color{limegreen}Guitar for beginners}. }  \\ 
\cline{1-3}
\textbf{Gameplay} & 301/124 & \makecell[l]{\emph{Q: What card did my opponent play before I placed the} \\ \emph{red {\color{limegreen}9}?}  A: Red {\color{limegreen}8}. }   \\ 
\cline{1-3}
\textbf{Others} & 201/88 & \makecell[l]{\emph{Q: Where might I find the office rules? } \\ A: On the notices on the wall near the entrance. } \\
\Xhline{1pt}
\end{tabular}
\vspace{-0.5cm}
\end{table}

\noindent\textbf{Raw Video Filtering} 
Our videos are drawn from two public ego-view video datasets, RoadTextVQA~\cite{tom2023reading} and EgoSchema~\cite{mangalam2023egoschema}. They cover outdoor and indoor scenarios, centered around driving and housekeeping activities. 
RoadTextVQA has 3,222 ten-second videos from dashcam footage of people driving. 
EgoSchema features 5,063 three-minute videos sourced from Ego4D~\cite{grauman2022ego4d}; its videos are mostly about indoor activities such as cooking, game playing, manufacturing, \etc.
To find videos rich in scene text, we first apply a state-of-the-art scene-text detection system~\cite{he2024gomatching} to the raw videos and threshold for videos with a significant percentage of frames with scene text: exceeding 5\% for Ego4D and 15\% for RoadTextVQA (since it features more scene text). 
After automatic filtering, we obtain 700 videos from RoadTextVQA~\cite{tom2023reading} and 1,800 from EgoSchema~\cite{grauman2022ego4d}. Since Ego4D videos are not collected for scene text study, we further filter the unqualified videos, mainly removing the tedious videos and videos that contain watermark or text present only on the clothes of the camera wearers. Here, we obtain 933/700 videos from Ego4D/RoadTextVQA.

\noindent\textbf{QA Generation}
Manually collecting QA pairs is labor-intensive; the QAs may also lack diversity. We use advanced MLLMs (\ie, GPT-4o) and generate the initial QAs with the following protocol: First, a video is decoded into 6 frames per second (fps) and evenly divided into 5 segments, wherein the frames without scene text are removed. Then, we uniformly sample 3 frames from each segment, and feed them to GPT-4o along with the prompts (in Appendix~{\color{red}D})~
to generate 3 QA pairs. This leads to a maximum of 15 QA pairs per video. The prompts are carefully designed to elicit questions that are:  
\textbf{(1)} goal-oriented, capturing real user needs by engaging with the visual scene from a first-person perspective;  
\textbf{(2)} aware of scene text, though not necessarily requiring the exact text transcription;  
\textbf{(3)} naturally expressed in a colloquial, first-person manner with referring expressions; and  
\textbf{(4)} challenging, as they demand an understanding of the video beyond a single image.

\begin{figure*}[t!]
\centering
\includegraphics[width=\textwidth]{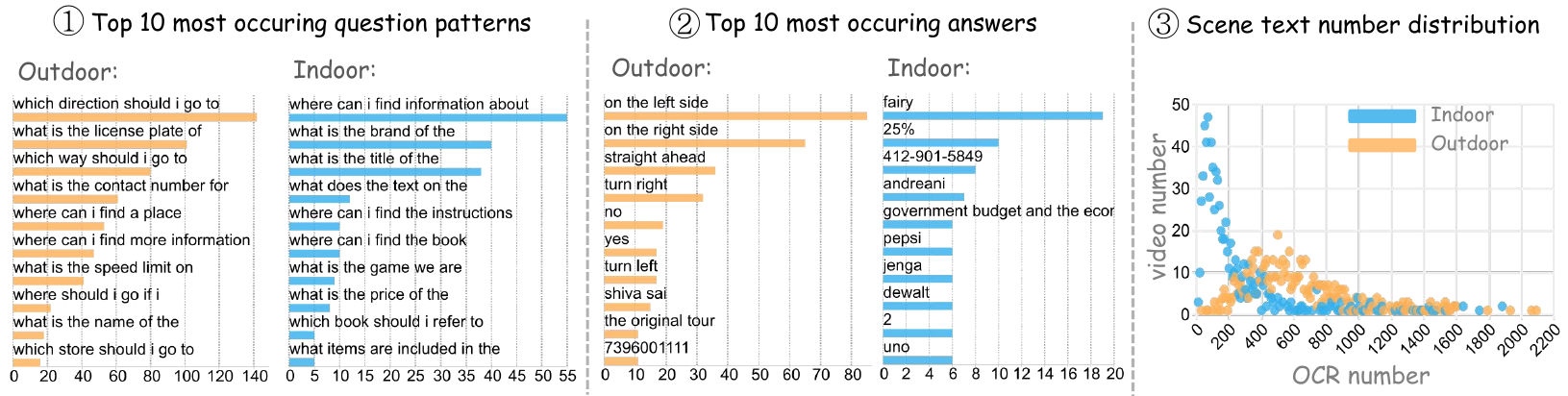}
\vspace{-0.5cm}
\caption{Distribution of QAs and OCR numbers.} 
\label{fig:data_dis}
\vspace{-0.2cm}
\end{figure*}

\begin{table*}[t!]
\centering
\caption{Comparison of related egocentric VQA and scene-text VQA test benchmarks. ST: Scene Text. STQ/STA: Questions/Answers contain scene text. EgoV: Egocentric Video. IntQ: Intentional Questions. QC: Question Category. OE/MC: Open-ended/Multi-choice.
}
\vspace{-0.2cm}
\label{tab1:dataset_compare}
\setlength{\tabcolsep}{.6em}
\centering
\fontsize{8}{10}\selectfont
\small
\resizebox{\linewidth}{!}{
\begin{threeparttable}
\begin{tabular}{lccccccccccccc}
\Xhline{1pt}
\textbf{Benchmark}  & \textbf{\#Que.} & \textbf{\#Video} & \textbf{Ave. Len. (s)} & \textbf{ST} & \textbf{STQ (\%)} & \textbf{STA (\%)} & \textbf{EgoV} & {\bf Indoor} & {\bf Outdoor} &  \textbf{Real Time} & \textbf{IntQ}  & \textbf{QC}  & \bf Task \\ 
\Xhline{1pt}
\multicolumn{3}{l}{\emph{Egocentric VQA Benchmarks}} \\
\hline
EgoThink~\cite{cheng2024egothink} & 700  & -  & - & \xmark  & - & - & \cmark &\cmark & \cmark  & \xmark & \xmark & \cmark & OE  \\
QAEgo4D~\cite{Baermann_2022_CVPR}  & 1,854 & 166  & 495.1 & \xmark & - & - & \cmark & \cmark & \cmark &  \xmark  & \xmark  & \xmark  & OE \\ 
AssistQ~\cite{wong2022assistq}  &  531 & 100 & 115.0 & \xmark & -  &  - & \cmark & \cmark & \xmark & \xmark &  \cmark & \xmark   & MC \\ 
EgoSchema~\cite{mangalam2023egoschema} & 5,063  & 5,063  & 180.0 & \xmark  &  - & -  & \cmark & \cmark & \cmark  & \xmark  &  \xmark  &  \xmark & MC \\  
EgoMemoria~\cite{ye2024mm} & 7,026 & 629 & 858.5 & \xmark &  - & -  & \cmark  & \cmark & \cmark & \xmark & \xmark & \cmark  & MC \\ 
\hline
\multicolumn{4}{l}{\emph{Scene-Text VQA Benchmarks}} \\
\hline
TextVQA~\cite{singh2019towards} & 5,734 & -  & - & \cmark & 16 & 85 & \xmark & \cmark & \cmark & \xmark & \xmark & \xmark  & OE \\ 
ST-VQA~\cite{biten2019scene}   & 4,163 & -  & - & \cmark & 18 & 92 & \xmark & \cmark &\cmark & \xmark & \xmark & \xmark  & OE \\ 
ESTVQA~\cite{wang2020general} & 5,014 & -  & - & \cmark  & 25 & 97 & \xmark &\cmark & \cmark & \xmark & \xmark & \xmark  & OE \\
M4ViteVQA~\cite{zhao2022towards}  & 2,103 & 680 & 5.7 & \cmark &  30 & 96 & \xmark & \cmark & \cmark & \xmark & \xmark & \xmark  & OE \\ 
RoadTextVQA~\cite{tom2023reading}  & 1,052 & 329 & 10.0  & \cmark & 60 & 65 & \cmark & \xmark & \cmark & \xmark & \xmark & \xmark  & OE \\ 
\rowcolor{gray!20} \dataset~(Ours)  & 7,064 & 1,507 &  101.7 & \cmark & 52 & 45 & \cmark  & \cmark & \cmark & \cmark & \cmark & \cmark & OE \\ 
\Xhline{1pt}
\end{tabular}
\vspace{-0.5cm}
\end{threeparttable}}
\end{table*}

\noindent\textbf{Manual Participation} 
\label{dataset:mp}
Despite using well-designed prompts, the generated QA pairs may fail to meet our strict requirements. Some generated questions are hallucinated or irrelevant to the visual content, have wrong formats, or are gibberish, \eg, involving special typeset characters not part of spoken language. We first employ GPT-4o to answer the questions without videos and remove the questions that can be answered blindly ($\sim$10\%). 
We then conduct manual filtering and correction for the remaining QAs. We provide some representative generated QA errors in Appendix~{\color{red}A}.

Specifically, we invite 9 well-trained university students for annotation. The manual check has 5 stages: (1) five annotators review for problematic QAs which may be redundant, not reflect real user needs, and irrelevant to scene text. The questions are amended where possible, or removed, leaving half of the original QAs; (2) and (3) two additional annotators re-examine and correct any mechanical issues, further reducing the QAs to 30\%; (4) five annotators refine the data to improve its colloquial quality; (5) four annotators strictly refine the remaining QAs according to the criteria specified in QA generation. The participants at each stage are asked to first check the remaining QAs from the prior stage. Additionally, we manually enrich $\sim$10\% time-sensitive questions to simulate real-time live QA by copying each suitable question to a different timestamp and invite five additional annotators to provide concise answers based on the visual content at the new timestamp.
Ultimately, we obtain 7,064 QA pairs of 1,507 videos to form the \dataset~dataset. Detailed question type and video scenario labels are provided. Figure~\ref{fig:exp} shows some representative examples; more are presented in Appendix~{\color{red}B.5}.

\subsection{Dataset Analysis}
\noindent\textbf{Dataset Statistics}
\dataset ~characterizes assistance-seeking questions related to scene text in diverse real-life scenarios. It consists of two parts: \textbf{(1) \datasetout}~focuses on the outdoor scenarios, with 694 videos and 4,848 QA pairs that may arise when driving; \textbf{(2) \datasetin~} emphasizes indoor scenarios, with 813 videos and 2,216 QA pairs that users may encounter in house-keeping activities. Distributions of videos and questions are listed in Table~\ref{tab:data_dis}. We also analyze the most frequent question and answer patterns in Figure~\cref{fig:data_dis} \textcircled{1}, \textcircled{2}. The statistics reveal a clear discrepancy between QAs; the driving questions are mostly related to navigation which pure scene text cannot answer, while the house-keeping questions are information seeking, where answers can be extracted from scene text. Figure~\ref{fig:data_dis} \textcircled{3} shows that the 
\textbf{\datasetin}~videos have a long-tailed distribution in the {number of OCR tokens}, with the majority having less than 200.  In contrast, \textbf{\datasetout}~videos are more evenly distributed with around 500.

\noindent\textbf{Dataset Comparison}
We compare \dataset~with several related benchmarks in Table~\ref{tab1:dataset_compare}. \dataset~has several unique features. Specifically, \dataset~is the first VideoQA testbed designed for \emph{egocentric scene-text aware QA assistance} in real-world scenarios, containing 7K QA pairs across 1.5K egocentric visual scenarios, covering both indoor and outdoor activities. The QA pairs focus on scene text comprehension, with about 45\% of answers and 52\% of questions referencing the exact scene text. Notably, \dataset~introduces \emph{real-time QA}, providing detailed timestamps for each question. While EgoMemoria~\cite{ye2024mm} also includes timestamps, its answers are based on the entire video, whereas \dataset~answers are derived from video content before the question timestamp. Additionally, the questions in \dataset~focus on inferring user intentions rather than pure visual understanding, with question categories provided for model analysis.

%% file: sec/4_experiment.tex
\section{Experiments}
\label{sec:expr}
\subsection{Experimental Setups}
\begin{table*}[t]
\caption{Performance of SoTA MLLMs on \datasetout. \textbf{Res.} denotes image resolution. We uniformly sample \textbf{\#F} frames that the models can accept from only video content before the timestamp. In this table, we test the performance of the model on low-resolution videos (640$\times$360 and 960$\times$540) of the \datasetout~ dataset. The human study is conducted on a random 30\% of all questions, with the only constraint that answers rely on video content before the question timestamp. The \textbf{best} and \underline{second-best} results are highlighted.}
\label{tab:expr_tab1}
\vspace{-0.2cm}
\setlength{\tabcolsep}{.6em}
\centering
\fontsize{8}{9}\selectfont
\resizebox{\textwidth}{!}{
\begin{threeparttable}
\begin{tabular}{l|c|c|c|ccccc|c}
\Xhline{1pt}  
{\textbf{Method}} &{\textbf{LLM}} & {\textbf{Res.}} & {\textbf{\#F}} & {\textbf{Location}} & {\textbf{Direction}}  & {\textbf{Description}} & {\textbf{Int. Reasoning}} & {\textbf{Others}} & {\textbf{Average}} \\ 
\Xhline{1pt}
\rowcolor{gray!20}
Human & - & - &  - & 38.7 / 2.3 & 32.4 / 2.2 & 54.9 / 3.0 & 45.3 / 2.6 & 36.9 / 2.5 & 43.1 / 2.6 \\ 
\hline
\multicolumn{10}{l}{\emph{\textcolor{lightgray}{Open-source Models}}} \\
ShareGPT4Video~\cite{chen2024sharegpt4video} & LLaMA3-8B & - & 60  & 11.7 / 1.5  & 20.8 / 1.8 & 5.5 / 0.7 & 10.7 / 1.3 & 10.0 / 1.2 & 11.5 / 1.3  \\
CogVLM2-Video~\cite{hong2024cogvlm2} & LLaMA3-8B & - & 60 & 10.6 / 1.5 & 15.7 / 1.7  & 12.4 / 1.0 & 12.4 / 1.3 & 10.0 / 1.3 & 12.4 / 1.4 \\ 
MiniCPM-V 2.6~\cite{yao2024minicpm} & Qwen2-7B & $448^{2}$ & 60 & 7.7 / 0.5 & 14.4 / 0.9 & 15.1 / 0.9 & 8.7 / 0.6 & 10.0 / 0.8 &  11.4 / 0.7 \\
VILA1.5~\cite{lin2024vila} & LLaMA3-8B & $384^{2}$ & 32 & 14.6 / 1.5  & 20.2 / 1.6 & 18.2 / {1.3} & 18.3 / {1.5} & 14.7 / {1.5} & 17.4 / {1.5}  \\ 
LLaVA-NeXT-Video~\cite{zhang2024llavanextvideo} & Qwen2-7B &  $384^{2}$ & 60 & {18.1} / 1.3 & {22.5} / 1.5 & {19.6} / 1.3 & {19.3} / 1.4 & {15.7} / 1.4  & {19.5} / 1.4 \\ 
InternVL2-8B~\cite{chen2024far} & InternLM2.5-7B & $448^{2}$  & 32  & 15.8 / 1.4  &  21.9 / 1.7 & 14.8 / 1.0 & 14.5 / 1.2  & 13.6 / 1.3  & 16.4 / 1.3 \\
InternVL2-26B~\cite{chen2024far} & InternLM2-20B & $448^{2}$  & 16  & 21.5 / 1.7 & 27.5 / \underline{2.0}  & 23.5 / 1.5  & 22.9 / 1.7  & 22.5 / 1.7 & 23.5 / 1.7   \\ 
Qwen2-VL~\cite{wang2024qwen2} & Qwen2-7B & -  & 16  & 23.4 / \underline{1.8} & \textbf{35.2} / \textbf{2.4}  & 30.8 / \underline{1.9} & 25.6 / \underline{1.9} & 22.5 / 1.8  & 28.2 / \textbf{2.0} \\ 
\hline 
\multicolumn{10}{l}{{\emph{\textcolor{lightgray}{Closed-source Models}}}} \\
GPT-4o~\cite{achiam2023gpt}  & - & $768^{2}$ & 32  & \underline{28.3} / {1.6}  & 24.9 / 1.4  & 35.0 / \underline{1.9}  & \underline{31.6} / 1.8 & \underline{29.8} / 1.8 & \underline{30.3} / 1.7   \\ 
Gemini 1.5 Flash~\cite{reid2024gemini} & - & $768^{2}$ & 32 & 25.4 / 1.6 & \underline{28.3} / {1.8} &  \underline{38.1} / \textbf{2.2} &  28.1 / {1.8} & 26.6 / \underline{2.0} &  30.1 / \underline{1.9} \\ 
Gemini 1.5 Pro~\cite{reid2024gemini}  & - & $768^{2}$ & 32  & \textbf{33.2} / \textbf{2.1} &  \underline{28.3} / {1.8} &  \textbf{38.8} / \textbf{2.2} & \textbf{32.7} / \textbf{2.0}  & \textbf{34.6} / \textbf{2.2}  &  \textbf{33.4} / \textbf{2.0}  \\ 
\Xhline{1pt}
\end{tabular}
\vspace{-0.5cm}
\end{threeparttable}}
\end{table*}

\noindent\textbf{Evaluation Metric}
We use GPT-4o mini \cite{gpt4omini2024} to evaluate the semantic similarity between the generated and the ground-truth (GT) answers, thus aligning closely with human scoring. Following \cite{maaz2023video}, we assess the generated answers with two metrics: Accuracy (0-100\%, the percentage of ``yes'' answers from the evaluator) and Score (0-5, with 5 being the highest match). Concretely, we format the question, predicted answer, and GT answer along with our customized evaluation prompts (in Appendix~{\color{red}D})~
for GPT-4o mini to determine the ``yes/no'' and score.

\noindent\textbf{Model Evaluation}
We evaluate three closed-source API-based models (GPT-4o~\cite{achiam2023gpt}, Gemini 1.5 Flash~\cite{reid2024gemini}, Gemini 1.5 Pro~\cite{reid2024gemini}) and seven advanced open-source MLLMs (MiniCPM-V 2.6~\cite{yao2024minicpm}, ShareGPT4Video~\cite{chen2024sharegpt4video}, InternVL2~\cite{chen2024far}, VILA1.5~\cite{lin2024vila}, LLaVA-NeXT-Video~\cite{zhang2024llavanextvideo}, CogVLM2-Video~\cite{hong2024cogvlm2}, Qwen2-VL~\cite{wang2024qwen2}) for evaluation. The models have reported the state-of-the-art (SOTA) on existing scene-text VQA benchmarks \cite{biten2019scene,singh2019towards,gurari2018vizwiz}, \eg, Qwen2-VL achieves QA accuracy $\sim$85\% on ST-VQA \cite{biten2019scene}. More details about these MLLMs and their specific prompts are presented in Appendix~{\color{red}B.1} and ~{\color{red}D}.

\subsection{Result Analysis}
\label{expr:ra}
Tables~\ref{tab:expr_tab1} and \ref{tab:expr_tab2} show that all models, especially the open-source ones, struggle to perform well on \dataset. The best results, achieved by Gemini~Pro~1.5~\cite{reid2024gemini}, are around 33$\sim$34\% on both \datasetout~and \datasetin, exceeding the best-performing open-source models by approximately $\sim$5\% and $\sim$9\% on \datasetout~and \datasetin, respectively. The performances vary significantly among open-source models, with Qwen2-VL~\cite{wang2024qwen2} and LLaVA-NeXT-Video~\cite{zhang2024llavanextvideo} performing the best, while MiniCPM-V 2.6~\cite{yao2024minicpm} and ShareGPT4Video~\cite{chen2024sharegpt4video} perform the worst.

\begin{table*}[t]
\caption{Performance of SoTA MLLMs on \datasetin. \textbf{Res.} denotes image resolution. We uniformly sample the \textbf{\#F} frames that the model can accept from all video frames before the timestamp as input. In this table, we test the performance of the model on the videos (the resolution mainly includes 480$\times$360 and 640$\times$360) of the \datasetin~ dataset. The human study is conducted on a randomly sampled 30\% of all questions. The best and second-best Accuracy / Score results are \textbf{bolded} and \underline{underlined} respectively.  }
\label{tab:expr_tab2}
\vspace{-0.2cm}
\setlength{\tabcolsep}{.3em}
\centering
\fontsize{7}{8}\selectfont
\resizebox{\textwidth}{!}{
\begin{threeparttable}
\begin{tabular}{l|c|c|c|cccccc|c}
\Xhline{1pt}  
{\textbf{Method}} &{\textbf{LLM}} & {\textbf{Res.}} & {\textbf{\#F}} & {\textbf{Hands-on}} & {\textbf{Kitchen}}  & {\textbf{Shopping}} & {\textbf{Gameplay}} & {\textbf{Book-Related}} & {\textbf{Others}} & {\textbf{Average}} \\ 
\Xhline{1pt}
\rowcolor{gray!20}
Human & - & - &  - & 26.6 / 1.9 & 31.3 / 2.1 & 18.2 / 1.4 & 34.8 / 2.2 & 27.2 / 1.8 & 32.8 / 2.1 & 27.7 / 1.9 \\ 
\hline
\multicolumn{11}{l}{\emph{\textcolor{lightgray}{Open-source Models}}} \\
ShareGPT4Video~\cite{chen2024sharegpt4video} & LLaMA3-8B & - & 128 & 7.0 / 1.0 & 5.1 / 1.0 & 5.9 / 1.3 & 12.3 / 1.2  & 7.1 / 1.1 & 4.0 / 0.9 & 7.0 / 1.1  \\
CogVLM2-Video~\cite{hong2024cogvlm2} & LLaMA3-8B & - & 128 & 12.3 / 1.4 & 11.6 / 1.4 & 10.9 / 1.4  & 23.6 / 1.8  & 11.2 / 1.2 & 13.4 / 1.4 & 13.5 / 1.4   \\ 
MiniCPM-V 2.6~\cite{yao2024minicpm} & Qwen2-7B & $448^{2}$ & 120 & 14.2 / 0.8 & 8.1 / 0.5  & 10.3 / 0.6 & 17.6 / 1.1  & 16.4 / 1.0 & 12.9 / 0.8 & 13.3 / 0.8   \\
VILA1.5~\cite{lin2024vila} & LLaMA3-8B & $384^{2}$ & 32 & 16.5 / 1.5 & 18.8 / 1.5 & 15.1 / 1.5  &  24.9 / 1.9 & 18.9 / 1.6 & 17.4 / 1.5 &  18.2 / 1.6  \\ 
InternVL2-8B~\cite{chen2024far} & InternLM2.5-7B & $448^{2}$  & 32  & 13.5 / 1.2 & 12.5 / 1.1 & 14.5 / 1.3 & 13.3 / 1.2  & 14.9 / 1.3  & 12.4 / 1.1 & 13.6 / 1.2 \\
InternVL2-26B~\cite{chen2024far} & InternLM2-20B & $448^{2}$  & 16  &  20.3 / 1.6 & 22.4 / 1.8  & 18.2 / 1.6  & 25.6 / 1.8 & 19.2 / 1.6 & 21.9 / 1.7 & 21.0 / 1.7    \\ 
Qwen2-VL~\cite{wang2024qwen2} & Qwen2-7B & $448^{2}$ & 48 & {21.5} / {1.8}  & {21.5} / {1.7}  & {22.5} / {1.8} & {29.9} / {2.1}  & {22.0} / {1.8}  & {26.4} / {1.8} & 23.3 / {1.8}    \\
LLaVA-NeXT-Video~\cite{zhang2024llavanextvideo} & Qwen2-7B &  $384^{2}$ & 128 & {24.4} / {1.7} & {23.6} / {1.7} & {21.8} / {1.7} & {29.9} / {1.9}  & {27.9} / {1.9}  & {27.8} / {1.8} & {25.4} / {1.8}  \\ 
\hline
\multicolumn{11}{l}{\emph{\textcolor{lightgray}{Closed-source Models}}} \\ 
GPT-4o~\cite{achiam2023gpt} & - & $768^{2}$ & 60 & 25.4 / 1.6 & 30.2 / 1.9  & 29.6 / 1.9  & 32.9 / 1.9  & 29.7 / 1.9 & 23.9 / 1.5 & 28.3 / 1.8   \\ 
Gemini 1.5 Flash~\cite{reid2024gemini} & - & $768^{2}$ & 60 & \underline{29.7} / \underline{2.0} & \underline{33.7} / \underline{2.1} & \underline{32.3} / \underline{2.1}  & \underline{34.5} / \underline{2.3}  & \underline{33.6} / \textbf{2.2}  & \underline{30.1} / \underline{2.0}  & \underline{32.0} / \underline{2.1}  \\ 
Gemini 1.5 Pro~\cite{reid2024gemini} & - & $768^{2}$ & 60  & \textbf{33.2} / \textbf{2.1} & \textbf{35.2} / \textbf{2.1}  & \textbf{33.8} / \textbf{2.1}  & \textbf{39.5} / \textbf{2.4}  & \textbf{34.3} / \underline{2.1} & \textbf{30.9} / \textbf{2.0} & \textbf{34.4} / \textbf{2.1}   \\ 
\Xhline{1pt}
\end{tabular}
\end{threeparttable}}
\vspace{-0.4cm}
\end{table*}


\begin{figure}[t!]
\centering
\includegraphics[width=1.0\columnwidth]{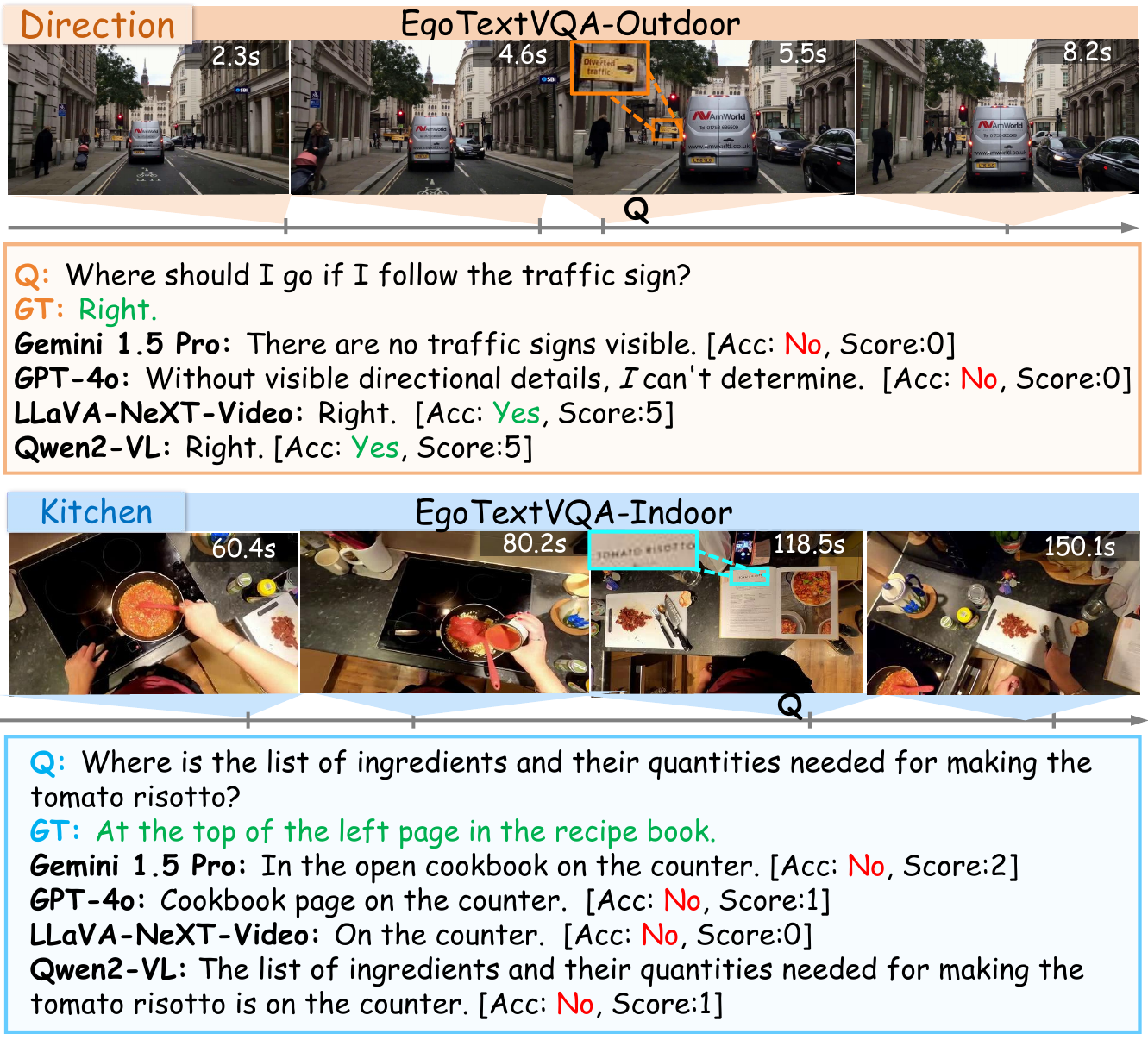}
\vspace{-0.2in}
\caption{Result Visualization.} 
\vspace{-0.6cm}
\label{fig:vis}
\end{figure}

\noindent\textbf{\datasetout} 
Table~\ref{tab:expr_tab1} shows that MLLMs struggle across different question categories. The open-source models perform particularly poorly on “Location” and “Intention Reasoning” questions, while closed-source models, GPT-4o and Gemini 1.5 Pro/Flash, often fail on “Direction” questions. “Location” questions require understanding spatial relationships, while “Intention Reasoning” involves inferring user intent from egocentric visual context. These challenges highlight the limitations of open-source models in fine-grained visual reasoning. For “Direction” questions, closed-source models frequently respond with phrases like “unknown scene text” suggesting difficulties in recognizing or locating specific scene text (see Figure~\ref{fig:vis} Top). Overall, Gemini 1.5 Pro~\cite{reid2024gemini} performs best, achieving 33.4\% accuracy and a score of 2.0, making it the SOTA model on \datasetout. However, it still trails human performance by $\sim$10\%, indicating that even advanced MLLMs struggle with scene text perception in complex, dynamic, egocentric environments.

\begin{figure}[t!]
\centering
\includegraphics[width=0.8\columnwidth]{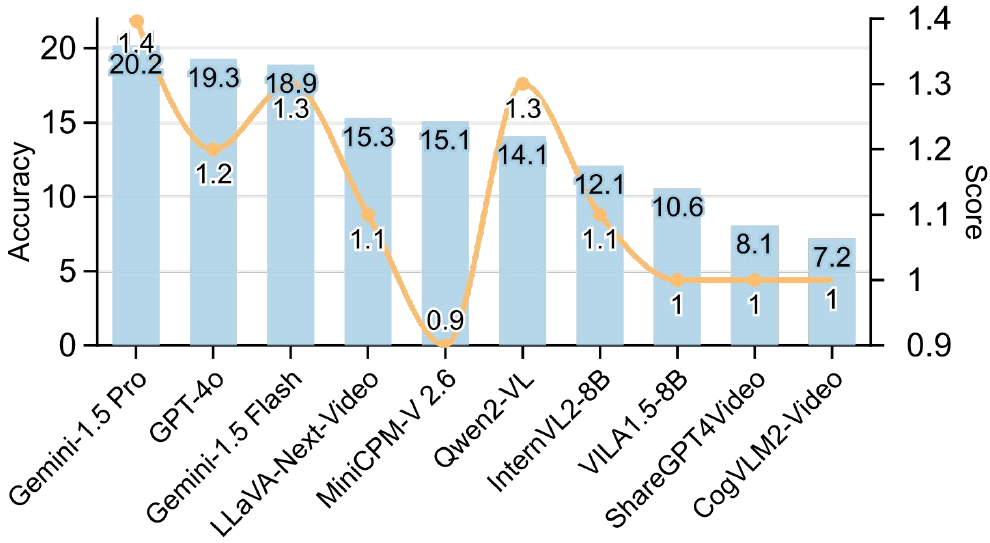}
\vspace{-0.2cm}
\caption{Performance of MLLMs on the real-time QA subset of \datasetout~($\sim$623 QA pairs).} 
\vspace{-0.4cm}
\label{fig:expr_2_hist}
\end{figure}

In Figure~\ref{fig:expr_2_hist}, we further analyze model behavior on the real-time QA subset of \datasetout. The results show that all models perform significantly worse on this subset, with the highest accuracy reaching only 20.2\% (\vs. 33.4\% on the full set), underscoring the substantial challenge of egocentric live QA. Interestingly, MiniCPM-V~2.6~\cite{yao2024minicpm} ranks among the top two open-source models for real-time QA, despite being the worst on the full set. However, a closer examination of its predictions reveals that MiniCPM-V~2.6 primarily excels at answering unanswerable questions with responses like ``I did not find it” whereas other models tend to hallucinate answers. This finding suggests that current models remain fundamentally weak in generating meaningful answers that satisfy user needs.

\noindent\textbf{\datasetin} 
Table~\ref{tab:expr_tab2} shows that MLLMs struggle with the questions across indoor scenarios, especially in “Kitchen” and “Shopping” activities. Our qualitative analysis (see Figure~\ref{fig:vis} Bottom) shows that people often ask about the location of ingredients and utensils in the kitchen, where the models tend to provide vague positional information (\eg, ``\texttt{on the left}" rather than ``\texttt{on the top shelf on the left}"). In shopping, it is challenging for the models to locate a user-specified item among large and densely packed products. ShareGPT4Video~\cite{chen2024sharegpt4video} performs the worst despite training on Ego4D videos, suggesting limited scene-text awareness. Meanwhile, closed-source models perform relatively better across scenarios.


Intriguingly, we find that humans perform even worse than the closed-source models. Feedback from our human annotators and a careful analysis of the answers suggest that the primary reasons are as follows: (1) scene text recognition is challenging for humans in Ego4D videos due to lower resolution, motion blur, occlusion, and long video; (2) answering the questions needs external knowledge beyond simple scene text recognition. \datasetout~mainly involves knowledge about store or restaurant chains (\eg, ``\texttt{Home Depo}" {sells building materials), and \datasetin~focuses on items like kitchen utensils, board games, and specialized books (\eg, identifying {a} ``\texttt{mandoline}" {slices potatoes} or {a} ``\texttt{quantum}" {{field theory book})};
(3) answer diversity also results in relatively lower human performance. 
The first two points underscore the challenges and significance of research on egocentric scene-text QA assistance. The third point unveils a limitation of our dataset, which we wish to solve in the future by enriching the GT answers for each question. We provide more human study analysis in Appendix~{\color{red}B.4} and ~{\color{red}C}.

\begin{table}[t!]
\caption{Effects of taking as input an image (``TS Frame") and three images (``QA Frames" analogous to QA generation) at the question timestamp.}
\label{tab:expr_tab3}
\setlength{\tabcolsep}{.1em}
\centering
\vspace{-0.2cm}
\fontsize{10}{12}\selectfont
\resizebox{\columnwidth}{!}{
\begin{threeparttable}
\begin{tabular}{l|c|ll|ll}
\Xhline{1pt}  
\multirow{2}{*}{\textbf{Method}} & \multirow{2}{*}{\textbf{Input}} & \multicolumn{2}{c|}{\textbf{\datasetout}} & \multicolumn{2}{c}{\textbf{\datasetin}}  \\
\cline{3-6}
& & Accuracy & Score & Accuracy & Score  \\
\Xhline{1pt}
\multirow{3}{*}{InternVL2-8B~\cite{chen2024far}} & Video & 16.4 & 1.3 & 13.6 & 1.2 \\ 
& TS Frame &  15.8 {\color{red} $\downarrow$ 0.6} &  1.2 {\color{red} $\downarrow$ 0.1} &  6.5 {\color{red} $\downarrow$ 7.1} &  0.7 {\color{red} $\downarrow$ 0.5} \\ 
& QA Frames &  18.5 {\color{limegreen} $\uparrow$ 2.1} &  1.4 {\color{limegreen} $\uparrow$ 0.1} &  17.7 {\color{limegreen} $\uparrow$ 4.1} &  1.4 {\color{limegreen} $\uparrow$ 0.2} \\ 
\hline
\multirow{3}{*}{MiniCPM-V 2.6~\cite{yao2024minicpm}} & Video & 11.4 & 0.7 & 13.3 & 0.8 \\ 
& TS Frame &  7.9 {\color{red} $\downarrow$ 3.5} &  0.2 {\color{red} $\downarrow$ 0.5} &  3.8 {\color{red} $\downarrow$ 9.5} &  0.3 {\color{red} $\downarrow$ 0.4} \\ 
& QA Frames &  {9.3} {\color{red} $\downarrow$ 2.1}  &  {0.6} {\color{red} $\downarrow$ 0.1} &  13.7 {\color{limegreen} $\uparrow$ 0.4} &  0.9 {\color{limegreen} $\uparrow$ 0.1} \\ 
\hline
\multirow{3}{*}{Qwen2-VL~\cite{wang2024qwen2}} & Video & {28.2} & {2.0} & {23.3} & {1.8} \\
& TS Frame & {30.9} {\color{limegreen} $\uparrow$ 2.7} & {2.1} {\color{limegreen} $\uparrow$ 0.1} &  15.7 {\color{red} $\downarrow$ 7.6} &  1.4 {\color{red} $\downarrow$ 0.4} \\
& QA Frames &  31.9 {\color{limegreen} $\uparrow$ 3.7} &  \textbf{2.2}  {\color{limegreen} $\uparrow$ 0.2} &  30.7 {\color{limegreen} $\uparrow$ 7.4} &  2.2 {\color{limegreen} $\uparrow$ 0.4} \\
\hline
\multirow{3}{*}{GPT-4o~\cite{achiam2023gpt}} & Video &  30.3 &  1.7 & 28.3  & 1.8   \\ 
& TS Frame &  27.7 {\color{red} $\downarrow$ 2.6} &   1.6 {\color{red} $\downarrow$ 0.1} &  14.8 {\color{red} $\downarrow$ 13.5} &  0.9 {\color{red} $\downarrow$ 0.9} \\ 
& QA Frames &  30.4 {\color{limegreen} $\uparrow$ 0.1}  &  1.8 {\color{limegreen} $\uparrow$ 0.1} &    \textbf{41.6} {\color{limegreen} $\uparrow$ 13.3} &   \textbf{2.4} {\color{limegreen} $\uparrow$ 0.6} \\ 
\hline
\multirow{3}{*}{Gemini 1.5 Pro~\cite{reid2024gemini}} & Video &  {\bf33.4} & {2.0}  &  34.4 & 2.1 \\ 
& TS Frame &  30.4 {\color{red} $\downarrow$ 3.0} & 1.8 {\color{red} $\downarrow$ 0.2} & 15.8 {\color{red} $\downarrow$ 18.6} & 1.1 {\color{red} $\downarrow$ 1.0} \\ 
& QA Frames & 26.3 {\color{red} $\downarrow$ 7.1} & 1.6 {\color{red} $\downarrow$ 0.4} & {38.2} {\color{limegreen} $\uparrow$ 3.8} & {2.4} {\color{limegreen} $\uparrow$ 0.3} \\ 
\Xhline{1pt}
\end{tabular}
\vspace{-0.6cm}
\end{threeparttable}}
\end{table}

\noindent\textbf{Outdoor Driving \vs Indoor House-Keeping} 
While humans perform worse on \datasetin, the MLLMs are basically at the same performance level on both datasets. In particular, we find Qwen2-VL~\cite{wang2024qwen2} stands out as the best open-source model on \datasetout, whereas LLaVA-NeXT-Video~\cite{zhang2024llavanextvideo} ranks the top on \datasetin. We speculate that LLaVA-NeXT-Video is better at coping with long videos. More model design and case analysis are provided in Appendix~{\color{red}B.2} and~{\color{red}B.5}.



\vspace{-0.2cm}
\section{Heuristic Solution Investigations}
To facilitate future study, we conduct extensive experiments to thoroughly analyze the challenges presented by \dataset~in Tables~\ref{tab:expr_tab3} to \ref{tab:expr_modal} and Figures \ref{fig:expr_2_hist} to \ref{fig:expr_1_leida}.  Our analysis centers on the following four questions: 

\noindent\textbf{Q1: Can image-level understanding solve our task?  Sometimes.}
We test the models by inputting a single frame at the question timestamp (``TS Frame"). Compared with using uniformly sampled video frames, Table~\ref{tab:expr_tab3} shows that almost all the models have a significant performance drop ranging from 0.6$\sim$3.5\% on \datasetout ~and as much as 7.1$\sim$18.6\% on \datasetin. This signifies the importance of video-level reasoning, especially for \datasetin, where questions feature longer video action understanding. Notably, Qwen2-VL \emph{improves} its performance when fed with a single frame on \datasetout. We speculate that 
Qwen2-VL excels at scene text recognition at static frames but shows insufficient capability in handling information redundancy across multiple frames.
\begin{figure}[t!]
\centering
\includegraphics[width=\columnwidth]{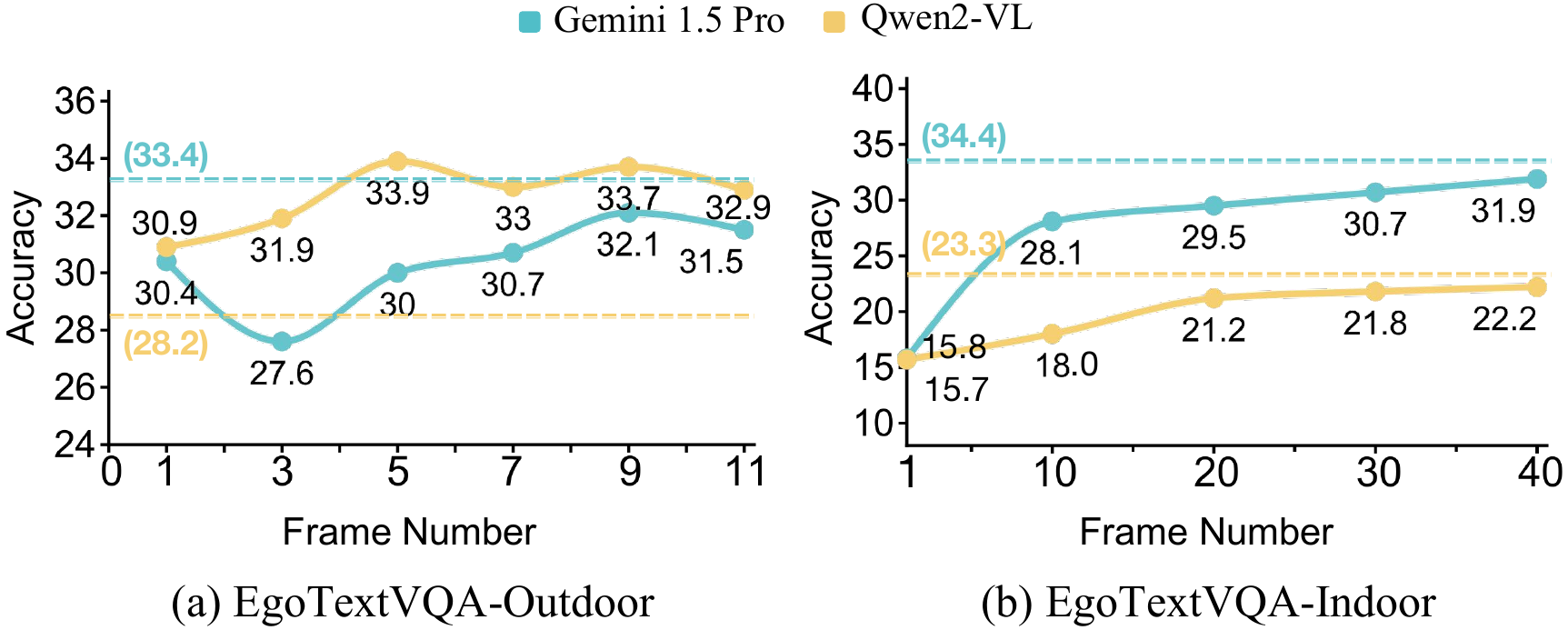}
\vspace{-0.2in}
\caption{Effects with different numbers of frame inputs sampled at 1fps backward from the question timestamp.} 
\label{fig:expr_3_zhexian}
\vspace{-0.6cm}
\end{figure}

\noindent\textbf{Q2: Is temporal grounding important? Sometimes.}
We use the three frames from question generation as model inputs (``QA Frames"), which consistently improve performance on \datasetin~but have inconsistent effects on \datasetout in Table~\ref{tab:expr_tab3}.
Qwen2-VL~\cite{wang2024qwen2} and InternVL2-8B~\cite{chen2024far} increase their accuracy by 3.7\% and 2.1\% respectively but MiniCPM-V 2.6~\cite{yao2024minicpm} and Gemini 1.5 Pro decrease by 2.1\% and 7.1\%, respectively. Overall, the improvement suggests there is space for models to optimize their temporal grounding, \ie, to localize the key frames for answering.
In contrast, the decrease in performance indicates that the models may be vulnerable to seriously reduced information input. It may also reflect the great efforts made by annotators for correcting mistakes from the automatic QA generation, as also exemplified by the poor results of GPT-4o (QA generator).

We investigate a question-timestamp aware sampling strategy: starting from the question timestamp and sampling backwards at 1 fps. This strategy is inspired by intuition that people often pose questions based on their most recent visual settings. Figure \ref{fig:expr_3_zhexian} shows that incorporating such a sampling strategy significantly improves Qwen2-VL's performance on \datasetout, even with fewer frames. However, for Gemini 1.5 Pro, the opposite is true and more frames are key for better performance. On \datasetin, we find that both Qwen2-VL and Gemini 1.5 Pro shrink their performances compared with standard uniform sampling across the whole video, possibly due to a lack of necessary information for long video reasoning.

\begin{figure}[t!]
\centering
\includegraphics[width=\columnwidth]{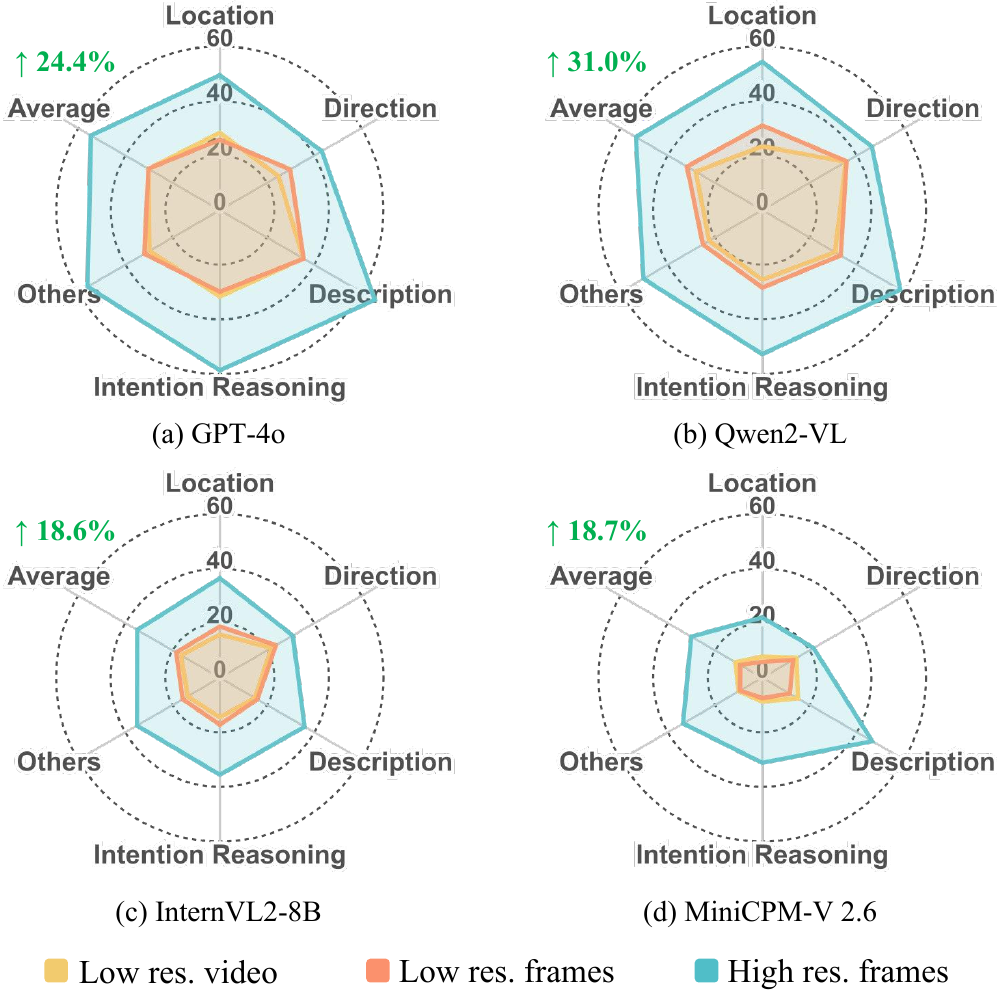}
\caption{Higher resolution (1920$\times$1080, 1280$\times$720) QA Frames generally improve performance on \datasetout. } 
\label{fig:expr_1_leida}
\vspace{-0.3cm}
\end{figure}

\noindent\textbf{Q3: Are high-resolution frames for scene text critical? Yes.} Most existing MLLMs are limited to accepting fixed, low-resolution images as input. When higher resolutions are used, the number of input frames is limited. We compare the performance of the model when using the “low-resolution video” and the three key “low-resolution frames” and “high-resolution frames” specified for question generation as input. Here we choose GPT-4o~\cite{achiam2023gpt} and three open-source models that can accept frames with resolutions of up to 1920$\times$1080 and 1280$\times$720. Figure~\ref{fig:expr_1_leida} shows that answering with three key frames (``QA Frames'') at low resolution is comparable to uniform video sampling. Yet, with the increase of resolution, GPT-4o~\cite{achiam2023gpt}, Qwen2-VL~\cite{wang2024qwen2}, InternVL2-8B~\cite{chen2024far}, and MiniCPM-V 2.6~\cite{yao2024minicpm} all show profound improvements, up to 18.6$\sim$31.0\%. 

Aside from increasing the \emph{global frame} resolution, we focus on increasing the \emph{local scene-text} resolution while keeping the global frame resolution unchanged. We use Microsoft Azure OCR\footnotetext{\href{https://azure.microsoft.com/en-us/services/cognitive-services/computer-vision/}{https://azure.microsoft.com/en-us/services/cognitive-services}} for scene text detection and apply the SOTA scene text super-resolution model DiffTSR~\cite{zhang2024diffusion} to increase the resolution of the detected regions to 380 $\times$ 128. To project the super-resolution scene text back to the original low-resolution video frames, we consider three options: resize to 1.00 $\times$, 1.25$\times$, and 1.50$\times$ of the original scene text size.
Table \ref{tab:str} shows that higher-resolution scene text improves performance on \datasetout~but yields inconsistent results on \datasetin. We speculate that scene text detection in Ego4D videos is more challenging than in RoadTextVQA due to lower resolution and complex perspectives. This is further supported by the smaller performance gains on \datasetin~(\vs.~Outdoor) with additional OCR results in Table \ref{tab:expr_modal}.

\begin{table}
\caption{Effects of scene-text resolution and area ratio. \emph{Scale} parameter is to control the magnification ratio of the scene text regions. SR: Super Resolution Scene Text. We experiment with 30\% of the data for efficiency.}
\label{tab:str}
\vspace{-0.2cm}
\setlength{\tabcolsep}{.4em}
\centering
\label{tab:expr_tab6}
\fontsize{10}{12}\selectfont
\resizebox{\columnwidth}{!}{
\begin{threeparttable}
\begin{tabular}{l|c|cc|cc}
\Xhline{1pt}  
\multirow{2}{*}{\textbf{Method}} & \multirow{2}{*}{\textbf{Scale}} & \multicolumn{2}{c|}{\textbf{\datasetout}} & \multicolumn{2}{c}{\textbf{\datasetin}}  \\
\cline{3-6}
& & Accuracy & Score & Accuracy & Score  \\
\hline
{Qwen2-VL~\cite{wang2024qwen2}} & - & {28.2} & {2.0} & {23.3} & {1.8} \\
\hline
\multirow{3}{*}{Qwen2-VL w/ SR} 
& 1.00 $\times$ & 31.9 & 2.1 &  24.2 & 1.8 \\
&  1.25 $\times$  &  33.1  &  2.2  &  23.5  & 1.8  \\
&  1.50 $\times$  &  34.1 & 2.2 &  22.3  & 1.7 \\
\hline
Gemini 1.5 Pro~\cite{reid2024gemini} & - & 33.4  & 2.0 & 34.4  & 2.1 \\
\hline
\multirow{3}{*}{Gemini 1.5 Pro w/ SR} 
& 1.00 $\times$  & 36.9  & 2.2 &  31.7 & 2.1 \\
& 1.25 $\times$ &  38.1  &  2.2  &  30.1  & 2.0  \\
& 1.50 $\times$ &  38.1 & 2.2 & 30.2 & 2.0   \\
\Xhline{1pt}
\end{tabular}
\end{threeparttable}}
\vspace{-0.3cm}
\end{table}

\begin{table}[t!]
\caption{Effects of different modality inputs. V: Video. Q: Question. ST: Scene Text.}
\vspace{-0.2cm}
\label{tab:expr_modal}
\setlength{\tabcolsep}{.1em}
\centering
\fontsize{13}{15}\selectfont
\resizebox{\columnwidth}{!}{
\begin{threeparttable}
\begin{tabular}{l|ccc|ll|ll}
\Xhline{1pt}  
\multirow{2}{*}{\textbf{Method}} & \multicolumn{3}{c|}{\textbf{Input}} & \multicolumn{2}{c|}{\textbf{EgoTextVQA-Outdoor}} & \multicolumn{2}{c}{\textbf{EgoTextVQA-Indoor}} \\
\cline{2-8}
& V  & Q  & ST  & Accuracy  & Score & Accuracy  & Score  \\
\Xhline{1pt}
\multirow{4}{*}{Qwen2-VL~\cite{wang2024qwen2}} & - & \checkmark & - & 2.1  &  0.5 &  2.5  &  0.5  \\
& - & \checkmark & \checkmark & 15.4  & 1.2   &  13.3 & 1.1   \\
\cline{2-8}
& \checkmark & \checkmark & - &   28.2 & 2.0  &  23.3 & 1.8   \\
& \checkmark & \checkmark & \checkmark  &  39.9 {\color{limegreen} $\uparrow$ 11.7}  &  2.5 {\color{limegreen} $\uparrow$ 0.5}  &  23.6 {\color{limegreen} $\uparrow$ 0.3} &  1.8 {\color{limegreen} $\uparrow$ 0.0} \\
\hline
\multirow{4}{*}{LLaVA-Next-Video~\cite{zhang2024llavanextvideo}} & - & \checkmark & - &  8.5 & 1.1& 6.0 & 1.0  \\
& - & \checkmark & \checkmark &  20.6 & 1.4  &  15.6  & 1.2   \\
\cline{2-8}
& \checkmark & \checkmark & - & 19.5 & 1.4 & 22.7 & 1.8   \\
& \checkmark & \checkmark & \checkmark  & 37.4 {\color{limegreen} $\uparrow$ 17.9 } & 2.3 {\color{limegreen} $\uparrow$ 0.9 } & 30.8 {\color{limegreen} $\uparrow$ 8.1} & 2.1 {\color{limegreen} $\uparrow$ 0.3} \\
\hline
\multirow{4}{*}{GPT-4o~\cite{achiam2023gpt}}& - & \checkmark & - & 5.26 & 0.84 & 5.7 & 0.9  \\
& - & \checkmark & \checkmark & 24.1  & 1.71  & 17.6 & 1.4  \\
\cline{2-8}
& \checkmark & \checkmark & - & 30.3  & 1.7 & 28.3  & 1.8  \\
& \checkmark & \checkmark & \checkmark  & {\bf52.9} {\color{limegreen} $\uparrow$ 22.6} & {\bf3.0} {\color{limegreen} $\uparrow$ 1.3} & 37.9 {\color{limegreen} $\uparrow$ 9.6 } & 2.3 {\color{limegreen} $\uparrow$ 0.5 } \\
\hline
\multirow{4}{*}{Gemini 1.5 Pro~\cite{reid2024gemini}}& - & \checkmark & - & 4.3  &  0.8 & 4.6  & 0.8 \\
& - & \checkmark & \checkmark & 18.2 & 1.3  & 15.2  & 1.0   \\
\cline{2-8}
& \checkmark & \checkmark & - & 33.4 & 2.0 & 34.4 & 2.1  \\
& \checkmark & \checkmark & \checkmark  & 49.5 {\color{limegreen} $\uparrow$ 16.1 } & 2.9 {\color{limegreen} $\uparrow$ 0.9 } & {\bf39.5} {\color{limegreen} $\uparrow$ 5.1 } & {\bf2.4} {\color{limegreen} $\uparrow$ 0.3 } \\
\Xhline{1pt}
\end{tabular}
\end{threeparttable}}
\vspace{-0.5cm}
\end{table}

\noindent\textbf{Q4: Is text alone enough to answer questions? No.}
We feed the best-performing open-source and closed-source models with only the scene text, but as Table~\ref{tab:expr_modal} shows, none of these models can answer the questions reasonably. Pure scene text inputs are sufficient for answering some questions, but video input is more important for almost all the models. Furthermore, adding scene text together with video input further boosts the performances significantly, leading to the highest results in all our explorations. 
For instance, GPT-4o achieves 52.9\% on \datasetout ~and Gemini's accuracy on \datasetin~reaches to 39.5\%. The findings demonstrate the importance of both video and scene text inputs for better QA assistance.
More heuristic solution investigations can be found in Appendix~{\color{red}B.3}.

%% file: sec/5_conclusion.tex
\section{Conclusion}
To provide a benchmark for research on egocentric scene-text aware QA assistance, we carefully construct the \dataset~dataset. \dataset~highlights QA assistance involving scene text from an ego-perspective in diverse real-life scenarios, including outdoor driving and indoor house-keeping. The questions reflect the real user needs, yet the visual attention is often not focused on the scene text. With well-classified visual scenarios and question types, we comprehensively analyze models that excel on existing scene-text VQA benchmarks and find that they struggle on \dataset. We further explore heuristic solutions and provide insights for improvements,
With these efforts, we hope this work can complement existing VideoQA research in advancing real-life egocentric QA assistance.

%% file: sec/6_acknowledgements.tex
\section{Acknowledgements}

This research was funded by the NUS Artificial Intelligence Institute (NAII) seed grant number NAII-SF-2024-003. Additionally, this work was supported by the National Key R\&D Program of China (NO.2024YFB3311602),  the National Natural Science Foundation of China (72188101, 62020106007, 62272435), the Major Project of Anhui Province (202203a05020011), the Anhui Provincial Natural Science Foundation (2408085J040), the Fundamental Research Funds for the Central Universities (JZ2023YQTD0072, JZ2024HGTG0309, and JZ2024AHST0337), and the New Cornerstone Science Foundation through the XPLORER PRIZE. We sincerely appreciate the dedication and meticulous efforts of our dataset annotators, whose contributions were invaluable in ensuring data quality and advancing this research.

%% file: sec/X_suppl.tex
\clearpage
\setcounter{page}{12}
\maketitlesupplementary

\setcounter{section}{0} 
\renewcommand{\thesection}{\Alph{section}}

\section{\dataset~Dataset} 
\label{sup:data}
\subsection{Manual Participation}
\begin{figure*}
\centering
\includegraphics[width=\textwidth]{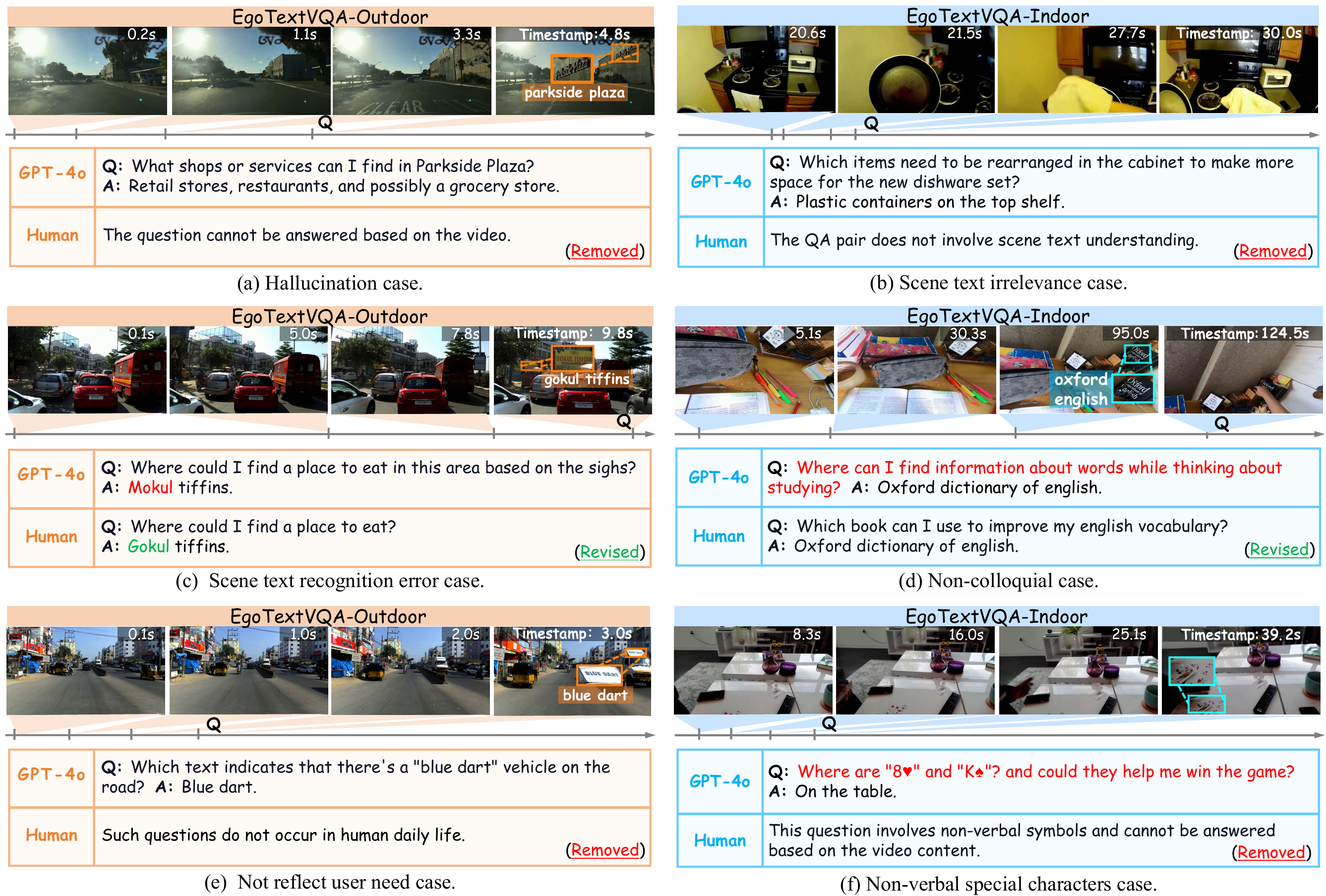}
\caption{Manual participation on \dataset~creation.}
\label{fig:supp_human}
\end{figure*}

We present examples of QA pairs generated by GPT-4o in Figure~\ref{fig:supp_human} to highlight the issues of automatic generation and underscore the value of manual correction. The primary problems observed are as follows: 
\textbf{(a) Hallucinated Answers:} The generated answers are unseen from the visual environment and cannot be confirmed by the annotators.
\textbf{(b) Scene Text Irrelevance:} The questions have vague references and fail to incorporate scene text understanding for answers. 
\textbf{(c) Scene Text Errors:} The questions or answers contain incorrect scene text. 
\textbf{(d) and (f) Non-colloquial Questions:} The questions are mechanical; they are phrased unnaturally and do not align well with daily spoken language. 
\textbf{(e) Not Reflect User Needs:} The question does not reflect real user needs and hardly occurs in human daily life.
After manual participation, about 70\% of the generated QAs are deleted and 30\% of the remaining QAs are revised.

\section{Experiment}
\subsection{Model Details}
\label{sup:models}
We provide a concise introduction to the MLLMs evaluated in Section~{\color{red}4}
, as outlined below:
\begin{itemize}
    \item \textbf{GPT-4o}~\cite{achiam2023gpt} advances the GPT-4 family towards more natural human-computer interactions. 
    \item \textbf{Gemini 1.5 Pro}~\cite{reid2024gemini} builds on Gemini 1.0’s~\cite{team2023gemini} research advances and multimodal capabilities and it is optimized for a wide-range of reasoning tasks. 
    \item \textbf{Gemini 1.5 Flash}~\cite{reid2024gemini} is a model from Gemini 1.5 family offering low latency and enhanced performance. 
    \item \textbf{Qwen2-VL}~\cite{wang2024qwen2} employs a ViT-675M~\cite{radford2021learning} as the visual encoder, Qwen2-7B as the language model, and an MLP projector. It improves upon Qwen-VL~\cite{bai2023qwen} with (1) naive dynamic resolution, allowing ViT to handle images of varying resolutions, and (2) multimodal rotary position embedding, which decomposes positional encoding into temporal, height, and width components. Qwen2-VL is pre-trained on diverse datasets, including image-text pairs, OCR data, interleaved articles, VQA datasets, video dialogues, and image knowledge sources, enabling a stronger multimodal understanding.
    \item \textbf{LLaVA-NeXT-Video}~\cite{zhang2024llavanextvideo} choose SigLIP-SO400M~\cite{zhai2023sigmoid} as the visual encoder, Qwen2~\cite{wang2024qwen2} as the language model, and a two-layer MLP as the projector. It utilizes the AnyRes~\cite{liu2024llavanext} technique to segment high-resolution images for the visual encoder and extends this approach to video processing. LLaVA-NeXT-Video has excellent reasoning, OCR, and world knowledge capabilities, achieving strong performance in video-based multimodal tasks.
    \item \textbf{VILA1.5}~\cite{lin2024vila} integrates CLIP-L~\cite{radford2021learning} as the visual encoder, LLaMA-2~\cite{touvron2023llama} as the language model, and a linear projector. It fine-tunes on a mix of internal data, including OCR-VQA~\cite{mishra2019ocr} and ST-VQA~\cite{biten2019scene}, and improves contextual learning by unfreezing the LLM during interleaved image-text pre-training. VILA1.5 excels in video reasoning, in-context learning, visual chain-of-thought reasoning, and world knowledge.
    \item \textbf{InternVL2-8B}~\cite{chen2024far} integrates InternViT-300M~\cite{chen2024internvl} with InternLM2.5-7B~\cite{cai2024internlm2} via a randomly initialized MLP projector. It is trained on OCR datasets generated by PaddleOCR~\cite{li2022pp}, utilizing Chinese images from Wukong and English images from LaionCOCO~\cite{schuhmann2021laion}. Building on the strong visual representations and high-resolution image processing capabilities of InternVL1.5~\cite{chen2024far}, InternVL2 incorporates instruction tuning, enabling competitive performance in document and chart comprehension, infographics QA, scene text understanding, OCR, and multimodal reasoning tasks.
    \item \textbf{CogVLM2-Video}~\cite{hong2024cogvlm2} utilizes the EVA-CLIP~\cite{sun2023eva} as the visual encoder, LLaMA3-8B as the language model, and a 2$\times$2 convolutional layer followed by a SwiGLU~\cite{shazeer2020glu} as the adapter. Unlike CogVLM~\cite{wang2023cogvlm}, CogVLM2 improves pre- and post-training data diversity and quality. The Synthetic OCR Dataset, a key pre-training resource, includes four OCR scenarios: (1) synthetic OCR images with text generated in Python, (2) real-world images with PaddleOCR~\cite{li2022pp}, (3) academic papers with extracted LaTeX via Nougat~\cite{blecher2023nougat}, and (4) HTML/LaTeX-rendered tables and formulae. CogVLM2-Video adapts CogVLM2 for videos, enhancing open-domain QA with temporal localization and timestamp-aware QA.
    \item \textbf{MiniCPM-V 2.6}~\cite{yao2024minicpm} employs SigLIP-SO400M~\cite{zhai2023sigmoid} as the visual encoder, Qwen2~\cite{wang2024qwen2} as the language model, and a compression module with one-layer cross-attention and a moderate number of queries as the projector. Its training includes pre-training on English and Chinese image captioning and OCR data, followed by fine-tuning on datasets like TextVQA~\cite{singh2019towards}, OCR-VQA~\cite{mishra2019ocr}, and ST-VQA~\cite{biten2019scene}. MiniCPM-V 2.6 excels in conversational and reasoning tasks across multiple images and videos, with high-resolution perception enabling features like table-to-markdown conversion and OCR transcription.
    \item \textbf{ShareGPT4Video}~\cite{chen2024sharegpt4video} builts on LLaVA-Next-8B~\cite{li2024llava}. Based on the proposed ShareGPT4Video~\cite{chen2024sharegpt4video} dataset, the proposed captioning model ShareCaptioner-Video generates high-quality captions with detailed temporal descriptions for various videos. ShareCaptioner-Video is fine-tuned with the collected video caption data. For video understanding, ShareGPT4Video's training dataset combines VQA samples from various instructional video-to-text datasets with video-caption pairs.
\end{itemize}

\subsection{Study of MLLM Design}
\label{sup:model_design}
We analyze key factors contributing to the superior performance of strong models (Qwen2-VL~\cite{wang2024qwen2}, LLaVA-NeXT-Video~\cite{zhang2024llavanextvideo}, and InternVL2~\cite{chen2024internvl}): 
(1) \textbf{enhanced visual encoder} capable of handling high-resolution and long-video inputs. In Table~\ref{tab:re_tab3}, increasing the number of video frames and resolution improves Qwen2-VL's performance by 1.2\% and 5.8\%;
(2) \textbf{more powerful LLM backbones}. Compared with InternVL2-8B, InternVL2-26B  performance has a 7\% increases in Table~\ref{tab:re_tab3}; and 
(3) \textbf{large-scale OCR training data}. Beyond the commonly used TextVQA datasets, InternVL2 leverages PaddleOCR to generate OCR samples for training. 
Additionally, we observe that as the number of video frames increases, Qwen2-VL's performance improves, whereas InternVL2's declines, underscoring the effectiveness of Qwen2-VL's video embedding design.

\subsection{Heuristic Solution Investigations}
\label{sup:heuristic}
\noindent\textbf{Effect of Timestamp-Aware Sampling}
We further investigate an alternative question-timestamp aware sampling strategy: starting from the question timestamp and uniformly sampling within fixed durations of 4 seconds and 32 seconds. As shown in Table~\ref{tab:supp_tab1}, on \datasetout, when sampling the same number of frames (\#F=16), this fixed duration sampling strategy achieves comparable or even superior performance to standard uniform sampling across the whole video for both Qwen2-VL~\cite{wang2024qwen2} and Gemini 1.5 Pro~\cite{reid2024gemini}. However, on \datasetin, when sampling \#F=48, we observe that while Qwen2-VL~\cite{wang2024qwen2} maintains comparable performance to standard uniform sampling, the performance of Gemini 1.5 Pro~\cite{reid2024gemini} drops by by about 7\%. This decline may stem from Gemini 1.5 Pro's stronger performance on questions requiring long-term video comprehension, which is less effectively captured by this fixed-duration sampling approach.

\noindent\textbf{Combination of Heuristic Strategies}
In the main text, we have explored different heuristic strategies separately. Here, we additionally study the combinations of heuristic strategies. First, for timestamp-aware video sampling, we adopt the strategy of ``\emph{fixed-duration sampling}'' to \datasetout~and ``\emph{1fps-backward sampling}'' to \datasetin, inspired by the results of the two different video sampling strategies on these two datasets.  
The results in Table~\ref{tab:supp_tab2} show that the models achieve cumulative performance improvements as heuristic strategies are progressively applied on \datasetout. Yet, the improvements are not stable on \datasetin, suggesting the significant challenge of egocentric scene-text aware QA assistance in daily house-keeping. 

\begin{table}
\caption{\small Study of MLLM design on \datasetout. The alignment module is MLP layer. VE: Visual Encoder.
}
\label{tab:re_tab3}
\setlength{\tabcolsep}{.4em}
\centering
\fontsize{8}{10}\selectfont
\resizebox{\columnwidth}{!}{
\begin{threeparttable}
\begin{tabular}{l|l|c|l|ll}
\hline 
\textbf{Method} & \textbf{VE} & \textbf{Res.} & \textbf{\#F} & Accuracy & Score  \\
\hline
\multirow{3}{*}{\makecell[c]{Qwen2-VL~\cite{wang2024qwen2} }}
& ViT-675M  & $448^{2}$ & 16 & 22.4 & 1.6 \\
& ViT-675M  & $448^{2}$ & 32 & 23.6 & 1.7 \\
& ViT-675M & - & 16 & 28.2 & 2.0 \\
\hline 
\multirow{2}{*}{InternVL2-8B~\cite{chen2024internvl}}  & InternViT-300M & $448^{2}$  & 16 & 16.5 & 1.3 \\
& InternViT-300M & $448^{2}$ & 32 & 16.4 & 1.3 \\
\hline 
InternVL2-26B~\cite{chen2024internvl}
& InternViT-6B & $448^{2}$ & 16 & 23.5 & 1.7 \\
\hline 
\end{tabular}
\end{threeparttable}}
\end{table}

\begin{table}
\caption{Effects of different numbers of video frames uniformly sampled within fix-duration before the question timestamp. We set a fixed duration of 4 seconds in \datasetout~and 32 seconds in \datasetin. S: Standard Uniformly Sampling. F: Fix-Duration Sampling. We experiment with 30\% of the data for efficiency.}
\label{tab:supp_tab1}
\setlength{\tabcolsep}{.4em}
\centering
\fontsize{8}{10}\selectfont
\resizebox{\columnwidth}{!}{
\begin{threeparttable}
\begin{tabular}{l|c|ll|c|ll}
\Xhline{1pt}  
\multirow{2}{*}{\textbf{Method}} & \multicolumn{3}{c|}{\textbf{\datasetout}} & \multicolumn{3}{c}{\textbf{\datasetin}}  \\
\cline{2-7}
& \#F & Accuracy & Score & \#F & Accuracy & Score  \\
\Xhline{1pt}
Qwen2-VL~\cite{wang2024qwen2} w/ S & 16 & 28.2 & 2.0  & 48  & 23.3 & 1.8  \\
\hline
\multirow{4}{*}{Qwen2-VL~\cite{wang2024qwen2} w/ F} 
& 4 & 26.1 & 1.8  & 12  & 16.2 & 1.4  \\ 
& 8 & 29.3 & 2.0  & 24  & 18.3 & 1.5   \\ 
& 12 & 29.9 & 2.0 & 36 & 22.1 & 1.7  \\ 
& 16 & 30.9 & 2.1 & 48 & 22.7 & 1.7  \\ 
\hline
Gemini 1.5 Pro~\cite{reid2024gemini} w/ S & 32 & 33.4 & 2.0  & 60  & 34.4 & 2.1  \\
\hline
\multirow{4}{*}{Gemini 1.5 Pro~\cite{reid2024gemini} w/ F}
& 4 & 27.9 & 1.7  & 12  & 22.8 & 1.6 \\ 
& 8 & 32.9 & 1.9  & 24  & 27.1 & 1.7 \\ 
& 12 & 33.5 & 2.0  & 36 & 29.0 & 1.8 \\ 
& 16 & 33.4 & 2.0  & 48 & 27.3 & 1.8  \\ 
\Xhline{1pt}
\end{tabular}
\vspace{-0.3cm}
\end{threeparttable}}
\end{table}

\subsection{Human Study}
\label{sup:human}
In Section~{\color{red}4}, the human results 
are based on two rounds of standard human studies. Based on the reason analysis for the poor human performance in Section~{\color{red}4.2},
we further validate the human performance by reducing the scene text recognition challenge. We sample 100 additional questions for humans to answer by providing the corresponding question frames. 
Table~\ref{tab:re_tab4} shows that humans perform better without the challenge of temporal grounding but still lag behind the best closed-source model (GPT-4o~\cite{achiam2023gpt}). This suggests advanced models may surpass humans in scene-text recognition or external knowledge, highlighting the importance of research on scene-text QA assistance.

\begin{table}
\caption{Effects of combining different heuristic strategies. T: Timestamp-Aware Sampling. ST: Additional Scene Text Input. HR: High-Resolution Scene Text (Scale = 1.25$\times$). We experiment with 30\% of the data for efficiency.}
\label{tab:supp_tab2}
\setlength{\tabcolsep}{.1em}
\centering
\fontsize{13}{15}\selectfont
\resizebox{\columnwidth}{!}{
\begin{threeparttable}
\begin{tabular}{l|ccc|cc|cc}
\Xhline{1pt}  
\multirow{2}{*}{\textbf{Method}} & \multicolumn{3}{c|}{\textbf{Input}} & \multicolumn{2}{c|}{\textbf{EgoTextVQA-Outdoor}} & \multicolumn{2}{c}{\textbf{EgoTextVQA-Indoor}} \\
\cline{2-8}
& T  & ST  & HR  & Accuracy  & Score & Accuracy  & Score  \\
\Xhline{1pt}
\multirow{4}{*}{Qwen2-VL~\cite{wang2024qwen2}} & - & - & - &  28.2 & 2.0 & 23.3  & 1.8 \\
& \checkmark &  -  &  - & 30.9 & 2.1 & 22.6 & 1.7 \\
& \checkmark & \checkmark & - & 42.4 & 2.7  & 25.3 & 1.8  \\
& \checkmark & \checkmark & \checkmark  & 42.6 & 2.7 & 27.1 & 1.9 \\
\hline
\multirow{4}{*}{Gemini 1.5 Pro~\cite{reid2024gemini}}& - & - & - & 33.4 & 2.0 & 34.4 & 2.1 \\
& \checkmark & - & - &  34.7 & 2.0  & 31.1 & 2.0  \\
& \checkmark & \checkmark & - &  49.5 & 2.9 & \bf38.0 & \bf2.3    \\
& \checkmark & \checkmark & \checkmark  & \bf51.1 & \bf3.0  & 36.5 & 2.2 \\
\Xhline{1pt}
\end{tabular}
\end{threeparttable}}
\end{table}

\begin{table}
\caption{Results of using video \emph{vs.} QA frames (three frames for QA generation) on \datasetin.}
\label{tab:re_tab4}
\setlength{\tabcolsep}{.6em}
\centering
\fontsize{6}{8}\selectfont
\resizebox{\columnwidth}{!}{
\begin{threeparttable}
\begin{tabular}{l|ll|ll}
\hline 
\multirow{2}{*}{\textbf{Method}}  & \multicolumn{2}{c|}{\textbf{Video}} & \multicolumn{2}{c}{\textbf{QA Frames}}  \\
\cline{2-5}
& Accuracy & Score & Accuracy & Score  \\
\hline
 Human  &  26.0  & 1.9  & 36.0 & 2.3   \\
\hline
GPT-4o~[27]   & 25.0  & 1.6  &  \bf39.0 &  \bf2.4  \\
Gemini-1.5~Pro~[29]   & \bf33.0  & \bf2.0 & 35.0 &  2.2  \\ 
\hline 
\end{tabular}
\end{threeparttable}}
\end{table}

\begin{figure*}
\centering
\includegraphics[width=\textwidth]{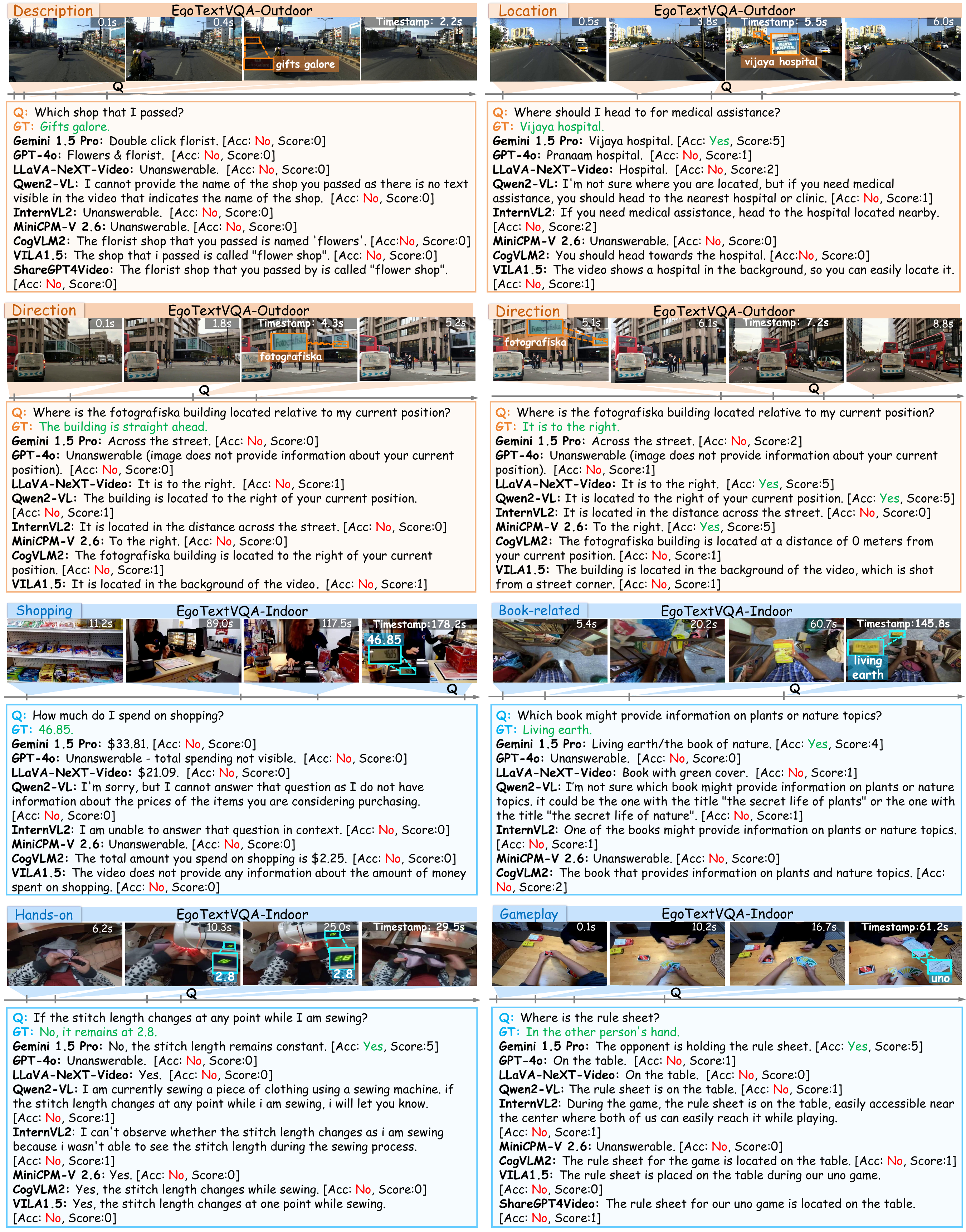}
\caption{Result visualization on \dataset.}
\label{fig:supp_results}
\end{figure*}

\subsection{Case Analysis}
\label{sup:qa_case}
As shown in Figure~\ref{fig:supp_results}, we qualitatively analyze the performance of MLLMs on \dataset. For \textbf{\datasetout}, the ``Description" example shows that all models struggle to accurately identify the target referred to by the question at the queried timestamp. The ``Location" example shows that only Gemini 1.5 Pro~\cite{reid2024gemini} correctly inferred the intention of the question and provided the precise answer. For the real-time ``Direction" examples, where the same question is posed at different timestamps, which corresponds to different answers, the left example shows that the questioned building is located right in front of the user at the question timestamp (4.3s), but all models fail to provide the correct answer, likely due to poor 3D spatial relation reasoning. 
Also, if the user poses the same question at 7.2s when he has moved to the left of the building, all models are unresponsive to such visual changes and tend to keep their original answers. This indicates that the models struggle to provide reasonable answers based on the real-time visual context in dynamic environment. 

For \textbf{\datasetin}, the ``Shopping" example shows that all models fail to effectively answer the total expense of 46.85 after the checkout process, indicating their
limitations in infer the total number after observing the price changes on the cashier's display during the checkout process. 
In the ``Book" example, the models need to identify the book from a large collection of books that matches the user’s needs (\eg, related to topics on plants or nature). The wrong answers indicate that the corresponding models are either weak at scene-text recognition and knowledge reasoning or tend to hallucinate their responses to match some question key words. Similar issues are also observed in the ``Hands-on" example.
Finally, the failures in the ``Gameplay'' example suggest that most models are weak in reasoning the real-time object state and people's real-time actions from an ego point of view. For example, while the ``rule sheet'' is on the table most of the time, it is on the other game player's hand at the time of user questioning.

\section{Agreement between Human and Evaluator}
\label{sup:agreement}
In this section, we evaluate the performance of models on \datasetout~using GPT-4o mini~\cite{gpt4omini2024} and human annotators. Following \cite{cheng2024egothink}, we invite three annotators to assess GPT-4o~\cite{achiam2023gpt} and Gemini 1.5 Pro~\cite{reid2024gemini}, the overall best-performing model. Human annotators maintain the same scoring principle as the model, as shown in Table~\ref{tab:supp_pro3}. We randomly sample 100 QA pairs for evaluation. As shown in Table~\ref{tab:re_tab1}, GPT-4o mini and human annotators achieve similar Accuracy and Score, with Pearson correlation coefficients of 0.80 and 0.87, respectively, indicating strong consistency. The Cohen's Kappa coefficients among three volunteers are 0.77 on Accuracy, indicating a high human agreement. 
To ensure reproducibility, future evaluations should use the same model version (GPT-4o-mini-2024-07-18) and the prompt in Table~\ref{tab:supp_pro3}.

\begin{table}[h]
\caption{\small Judgments of human and GPT-4o mini.}
\label{tab:re_tab1}
\setlength{\tabcolsep}{.4em}
\centering
\fontsize{6}{8}\selectfont
\resizebox{\columnwidth}{!}{
\begin{threeparttable}
\begin{tabular}{l|ll|ll}
\hline
\multirow{2}{*}{\textbf{Method}} & \multicolumn{2}{c|}{\textbf{GPT-4o~\cite{achiam2023gpt}}} & \multicolumn{2}{c}{\textbf{Gemini 1.5 Pro~\cite{reid2024gemini}}}  \\
\cline{2-5}
& Accuracy & Score & Accuracy & Score  \\
\hline
Human  & 36.0 & 1.9 & 47.3 & 2.5 \\
GPT-4o mini~\cite{gpt4omini2024}  & 34.0 & 1.8 & 42.0 & 2.3 \\
\hline
\end{tabular}
\end{threeparttable}}
\end{table}

\section{Model Prompts}
\label{sup:prompt}
Table~\ref{tab:supp_pro1} provides the prompts used by GPT-4o for question generation and filtering. Table~\ref{tab:supp_pro2} lists the prompts employed by GPT-4o for automatic question label annotation. Table~\ref{tab:supp_pro5} details the specific prompts applied for model inference. Table~\ref{tab:supp_pro3} shows the prompts used by GPT-4o-mini for model evaluation.  Table~\ref{tab:supp_pro4} includes the prompts designed for heuristic solutions with different modality inputs.

\begin{table*}
\caption{Prompts for question-answer generation and filtering on \dataset}
\centering
\label{tab:supp_pro1}
\fontsize{9}{10}\selectfont 
\begin{tabular}{p{17cm}}
\hline
\makecell[c]{\textbf{Question-Answer Generation Prompts}}  \\
\hline
\#\# Question Prompt: \\
Give you a first-person perspective video, which records the scene you see from the first-person perspective. Please judge from your perspective whether there are scene texts in the video. If so, please tell me what these scene texts are. Then, you have some questions about the scene texts you see, and ask three questions related to the activities you are going to carry out.
These scene texts can serve as clues to help you answer your questions.
Please generate three highly diverse questions based on the scene texts related to your activities in the first-person perspective video. 
If there is no scene text in the video, it is not necessary. Your questions should meet the following requirements:\\
Requirement 1: The questions should involve scene text understanding in the video.\\
Requirement 2: The questions should be goal-oriented and relevant to human daily life.\\
Requirement 3: The questions should require understanding multiple video frames, not just a single frame.\\
Requirement 4: The questions should be asked from a first-person perspective, expressed as colloquially as possible, and the first-person pronoun ``I" should be used appropriately.\\
Requirement 5: The questions should be of moderate length. \\
When announcing the question please label each question as ``Question 1, 2, 3: \{\emph{question}\}".\\
Please start your questions with the question word ``what", ``where", ``which", etc. You don't need to explain too much about what you are doing or indicate the location of the scene text in the video. Avoid the words ``video" and ``frame" in the questions. Remember to make sure that the correct answer to your question can be taken directly from the video and is concise enough.\\
Examples of good questions:\\
``Question 1: Which way is the exit?" \\
``Question 2: Could you tell me how much this item costs?" \\
``Question 3: What is the speed limit on this road?" \\
Image:\{\emph{image1}\}  Image:\{\emph{image2}\}  Image:\{\emph{image3}\} \\
\hline
\#\# Answer Prompt: \\
I provide three questions as follows: \{\emph{question}\} \\
You need to create an exam that tests above student abilities based on the three questions I just provided. Each question should have open-ended but short correct answers. Your answers have the following requirements:\\
Requirement 1: Your answers should be short and be closely related to the scene text in the video.\\
Requirement 2: Your answers should not mention any particular video frame number.\\
Requirement 3: Do not use letters for the answer choices.\\
You must print one correct answer and four wrong answers on separate lines in the following format: \"\ \\
Correct Answer :\{\emph{answer}\}\\
\hline
\makecell[c]{\textbf{Automatic Filtering Prompt}}  \\
\hline
You are a helpful assistant. You can answer the following questions based on your general knowledge. \\
Question: \{\emph{question}\} 
Answer briefly with a single word, a phrase, or a short sentence. \\
\hline
\end{tabular}
\end{table*}

\begin{table*}
\caption{Prompts for GPT-4o to annotate question categories on \datasetout.}
\centering
\label{tab:supp_pro2}
\fontsize{9}{11}\selectfont 
\begin{tabular}{p{17cm}}
\hline
\makecell[c]{\textbf{Question Classification Prompts}}  \\
\hline
Question: \{\emph{question}\} \\
Which of the following five categories does this question belong to? Please only answer the category name, such as Direction. \\
1. \textbf{Location}: Questions about a place or location. For example:  \\ 1) Where is the gas station? \\ 2) Which stores can I find on the right side of the road at this intersection? \\
2. \textbf{Direction}: Questions related to navigation, driving direction, and turns. For example:  \\ 1) Is the next road a left or right turn?  \\ 2) If I want to go to Cava, on which side of the street should I look for it?  \\ 3) Where should trucks go according to the signs? \\
3. \textbf{Description}: Questions that focus on scene text such as road signs, price labels, and billboards.  For example:  \\ 1) What does the sign on the side of the road say? \\ 2) What is the name of the center on the left side of the road? \\ 3) What is the name of the street to my right? \\
4. \textbf{Intention Reasoning}: Questions about behavioral activities involving drivers or passengers to solve personal needs.  For example: \\
1) Where do I need to go to solve my financial problems? \\ 2) Is there a place nearby where I can shop for appliances and electronics?  \\
5. \textbf{Others}: Composite questions that involve multiple different or the same types of the above, such as asking about both description and location.  For example: \\ 1) What event is being advertised on the bus, and where is it taking place? \\ 2) What is the contact number for the leadspace building, and what service might they provide? \\
\hline
\end{tabular}
\end{table*}

\begin{table*}
\caption{Prompts for MLLM inference on \dataset.}
\centering
\label{tab:supp_pro5}
\fontsize{9}{12.5}\selectfont 
\begin{tabular}{l|p{13.5cm}}
\hline
\textbf{Model} & \makecell[c]{\textbf{General Prompts}} \\
\hline
GPT-4o & Based on the following images from a video, please briefly answer the following question with a single word, a phrase, or a short sentence. Question: \{\emph{question}\}.  Output the answer to the question in the following format: Answer: \{\emph{answer}\}. If you cannot answer the question, please answer ``Unanswerable'' and briefly explain why you cannot answer. \\
\hline
Gemini 1.5 Flash & Based on the following images from a video, please briefly answer the following question with a single word, a phrase, or a short sentence. Question: \{\emph{question}\}.  Output the answer to the question in the following format: Answer: \{\emph{answer}\}. If you cannot answer the question, please answer ``Unanswerable'' and briefly explain why you cannot answer. \\
\hline
Gemini 1.5 Pro & Based on the following images from a video, please briefly answer the following question with a single word, a phrase, or a short sentence. Question: \{\emph{question}\}.  Output the answer to the question in the following format: Answer: \{\emph{answer}\}. If you cannot answer the question, please answer ``Unanswerable'' and briefly explain why you cannot answer. \\
\hline
LLaVA-Next-Video & Please answer the following questions related to this video.  If you cannot answer the question, please answer ``Unanswerable'' and briefly explain why you cannot answer. Keep your answer as short as possible. Keep your answer as short as possible. Keep your answer as short as possible. Question: \{\emph{question}\} \\
\hline
CogVLM2-Video & You are a person in the situation shown in the following consecutive images from a video. You can answer questions that humans ask to help them make decisions. Now you are observing your surroundings and answering questions based on the current situation. Understanding the scene text around you is important for answering questions.  Answer the questions in the first-person perspective. If you cannot answer the question, please answer ``Unanswerable'' and briefly explain why you cannot answer.  Question: \{\emph{question}\} \\ 
\hline
InternVL2 & You are a person in the situation shown in the following consecutive images from a video.  You can answer questions that humans ask to help them make decisions. Now you are observing your surroundings and answering questions based on the current situation. Understanding the scene text around you is important for answering questions. Answer the questions in the first-person perspective. If you cannot answer the question, please answer 'Unanswerable' and briefly explain why you cannot answer. Keep your answer as short as possible! Keep your answer as short as possible! Keep your answer as short as possible! Question: \{\emph{question}\}\\
\hline
Qwen2-VL &  You are a person in the situation shown in the following consecutive images from a video.  You can answer questions that humans ask to help them make decisions. Now you are observing your surroundings and answering questions based on the current situation. Understanding the scene text around you is important for answering questions. Answer the questions in the first-person perspective. If you cannot answer the question, please answer 'Unanswerable' and briefly explain why you cannot answer. Question: \{\emph{question}\} \\ 
\hline
VILA1.5 & You are a helpful language and vision assistant. You are able to understand the visual content that the user provides, and assist the user with a variety of tasks using natural language. Question: \{\emph{question}\} \\
\hline
ShareGPT4Video & You are a person in the situation shown in the following consecutive images from a video. You can answer questions that humans ask to help them make decisions. Now you are observing your surroundings and answering questions based on the current situation. Understanding the scene text around you is important for answering questions. Answer the questions in the first-person perspective. If you cannot answer the question. please answer ``Unanswerable" and briefly explain why you cannot answer. Question: \{\emph{question}\} \\
\hline
MiniCPM-V 2.6  & You are a person in the situation shown in the following consecutive images from a video.  You can answer questions that humans ask to help them make decisions. Now you are observing your surroundings and answering questions based on the current situation. Understanding the scene text around you is important for answering questions. Answer the questions in the first-person perspective. If you cannot answer the question, please answer 'Unanswerable' and briefly explain why you cannot answer. Keep your answer as short as possible! Keep your answer as short as possible! Keep your answer as short as possible! Question: \{\emph{question}\} \\
\hline
\end{tabular}
\end{table*}

\begin{table*}
\caption{Prompts for GPT-4o-mini to evaluate MLLMs on \dataset.}
\centering
\label{tab:supp_pro3}
\fontsize{9}{12}\selectfont
\begin{tabular}{p{17cm}}
\hline
\makecell[c]{\textbf{Evaluation Prompts}}  \\
\hline
You are an intelligent chatbot designed for evaluating the correctness of generative outputs for question-answer pairs.
Your task is to compare the predicted answer with the correct answer and determine if they match meaningfully. Here's how you can accomplish the task: \\
\\
\#\#INSTRUCTIONS: \\
- Focus on the meaningful match between the predicted answer and the correct answer. 
Please note that not only matches of noun phrases between answers, but also matches of prepositional phrases.  \\
For example, ``at the car wash on your right" does not exactly match ``car wash".  ``at the gas station beside the sign 'gas sale'" does not exactly match ``gas station"" \\
- Consider synonyms or paraphrases as valid matches. 
Note that the predicted answer must be consistent with the string type of the correct answer, which may include phone numbers, email addresses, numbers, dates, etc.  \\
For example, the string types ``www.usps.com" and ``visit their website" are inconsistent,  the string types ``9849041316" and ``advertiser's contact number" are inconsistent." \\
- Evaluate the correctness of the prediction compared to the answer." \\

\\
Please evaluate the following video-based question-answer pair: \\
Question: \{\emph{question}\} Correct Answer: \{\emph{GT answer}\}  Predicted Answer: \{\emph{predicted answer}\} \\
Provide your $eval_{code}$ only as a yes/no and score where the score is an integer value between 0 and 5, with 5 indicating the highest meaningful match. \\
Please generate the response in the form of a Python dictionary string with keys 'pred' and 'score', where the value of 'pred' is a string of 'yes' or 'no' and the value of 'score' is in INTEGER, not STRING.
DO NOT PROVIDE ANY OTHER OUTPUT TEXT OR EXPLANATION. Only provide the Python dictionary string. 
For example, your response should look like this: \{'pred': 'yes', 'score': 5\}, \{'pred': 'no', 'score': 1\}.\\
\hline
\end{tabular}
\end{table*}

\begin{table*}
\caption{Prompts for heuristic solution study of different modality inputs on \dataset.}
\centering
\label{tab:supp_pro4}
\fontsize{9}{12}\selectfont 
\begin{tabular}{l|p{14.5cm}}
\hline
\textbf{Model Input} & \makecell[c]{\textbf{Prompts}} \\
\hline
w/ Q & You are a helpful assistant. You can answer the following questions based on your general knowledge. Question: \{\emph{question}\}\\
\hline
w/ Q \& ST & You are a helpful assistant. You are provided with some important scene text information. You can answer the following questions based on your common sense or the scene text information I provide.  Please answer as briefly as possible. Please note that this scenario text information is very important. You can find the scene text related to the question as the answer. Scene Text: \{\emph{OCR results}\} Question: \{\emph{question}\} \\
\hline
w/ V \& Q \& ST & You are a person in the situation shown in the following consecutive images from a video. You can answer questions that humans ask to help them make decisions. Now you are observing your surroundings and answering questions based on the current situation. I will provide you with the following scene text that may be included in each image. Understanding the scene text is important for answering questions. Answer the questions in the first-person perspective. If you cannot answer the question, please answer 'Unanswerable' and briefly explain why you cannot answer. The scene texts in Frame {0} include: \{\emph{OCR results}\}. The scene texts in Frame 1 include: \{\emph{OCR results}\}. The scene texts in Frame {2} include: \{\emph{OCR results}\}. The scene texts in Frame \{\emph{frame id}\} include: \{\emph{OCR results}\}. Question: \{\emph{question}\}  \\
\hline
\end{tabular}
\end{table*}


%% file: main.bbl
\begin{thebibliography}{69}
\providecommand{\natexlab}[1]{#1}
\providecommand{\url}[1]{\texttt{#1}}
\expandafter\ifx\csname urlstyle\endcsname\relax
  \providecommand{\doi}[1]{doi: #1}\else
  \providecommand{\doi}{doi: \begingroup \urlstyle{rm}\Url}\fi

\bibitem[Bai et~al.(2023)Bai, Bai, Yang, Wang, Tan, Wang, Lin, Zhou, and Zhou]{bai2023qwen}
Jinze Bai, Shuai Bai, Shusheng Yang, Shijie Wang, Sinan Tan, Peng Wang, Junyang Lin, Chang Zhou, and Jingren Zhou.
\newblock Qwen-vl: A frontier large vision-language model with versatile abilities.
\newblock \emph{arXiv preprint arXiv:2308.12966}, 2023.

\bibitem[B\"armann and Waibel(2022)]{Baermann_2022_CVPR}
Leonard B\"armann and Alex Waibel.
\newblock Where did i leave my keys? - episodic-memory-based question answering on egocentric videos.
\newblock In \emph{CVPR Workshops}, pages 1560--1568, 2022.

\bibitem[Biten et~al.(2019)Biten, Tito, Mafla, Gomez, Rusinol, Valveny, Jawahar, and Karatzas]{biten2019scene}
Ali~Furkan Biten, Ruben Tito, Andres Mafla, Lluis Gomez, Mar{\c{c}}al Rusinol, Ernest Valveny, CV Jawahar, and Dimosthenis Karatzas.
\newblock Scene text visual question answering.
\newblock In \emph{ICCV}, pages 4291--4301, 2019.

\bibitem[Blecher et~al.(2023)Blecher, Cucurull, Scialom, and Stojnic]{blecher2023nougat}
Lukas Blecher, Guillem Cucurull, Thomas Scialom, and Robert Stojnic.
\newblock Nougat: Neural optical understanding for academic documents.
\newblock \emph{arXiv preprint arXiv:2308.13418}, 2023.

\bibitem[Cai et~al.(2024)Cai, Cao, Chen, Chen, Chen, Chen, Chen, Chen, Chen, Chu, et~al.]{cai2024internlm2}
Zheng Cai, Maosong Cao, Haojiong Chen, Kai Chen, Keyu Chen, Xin Chen, Xun Chen, Zehui Chen, Zhi Chen, Pei Chu, et~al.
\newblock Internlm2 technical report.
\newblock \emph{arXiv preprint arXiv:2403.17297}, 2024.

\bibitem[Chen et~al.(2024{\natexlab{a}})Chen, Wei, Li, Dong, Zhang, Zang, Chen, Duan, Lin, Tang, et~al.]{chen2024sharegpt4video}
Lin Chen, Xilin Wei, Jinsong Li, Xiaoyi Dong, Pan Zhang, Yuhang Zang, Zehui Chen, Haodong Duan, Bin Lin, Zhenyu Tang, et~al.
\newblock Sharegpt4video: Improving video understanding and generation with better captions.
\newblock \emph{arXiv preprint arXiv:2406.04325}, 2024{\natexlab{a}}.

\bibitem[Chen et~al.(2024{\natexlab{b}})Chen, Wang, Tian, Ye, Gao, Cui, Tong, Hu, Luo, Ma, et~al.]{chen2024far}
Zhe Chen, Weiyun Wang, Hao Tian, Shenglong Ye, Zhangwei Gao, Erfei Cui, Wenwen Tong, Kongzhi Hu, Jiapeng Luo, Zheng Ma, et~al.
\newblock How far are we to gpt-4v? closing the gap to commercial multimodal models with open-source suites.
\newblock \emph{arXiv preprint arXiv:2404.16821}, 2024{\natexlab{b}}.

\bibitem[Chen et~al.(2024{\natexlab{c}})Chen, Wu, Wang, Su, Chen, Xing, Zhong, Zhang, Zhu, Lu, et~al.]{chen2024internvl}
Zhe Chen, Jiannan Wu, Wenhai Wang, Weijie Su, Guo Chen, Sen Xing, Muyan Zhong, Qinglong Zhang, Xizhou Zhu, Lewei Lu, et~al.
\newblock Internvl: Scaling up vision foundation models and aligning for generic visual-linguistic tasks.
\newblock In \emph{CVPR}, pages 24185--24198, 2024{\natexlab{c}}.

\bibitem[Cheng et~al.(2024{\natexlab{a}})Cheng, Fang, Yu, Zhou, Li, Tian, Li, Han, and Liu]{cheng2024videgothink}
Sijie Cheng, Kechen Fang, Yangyang Yu, Sicheng Zhou, Bohao Li, Ye Tian, Tingguang Li, Lei Han, and Yang Liu.
\newblock Videgothink: Assessing egocentric video understanding capabilities for embodied ai.
\newblock \emph{arXiv preprint arXiv:2410.11623}, 2024{\natexlab{a}}.

\bibitem[Cheng et~al.(2024{\natexlab{b}})Cheng, Guo, Wu, Fang, Li, Liu, and Liu]{cheng2024egothink}
Sijie Cheng, Zhicheng Guo, Jingwen Wu, Kechen Fang, Peng Li, Huaping Liu, and Yang Liu.
\newblock Egothink: Evaluating first-person perspective thinking capability of vision-language models.
\newblock In \emph{CVPR}, pages 14291--14302, 2024{\natexlab{b}}.

\bibitem[Di and Xie(2024)]{di2024grounded}
Shangzhe Di and Weidi Xie.
\newblock Grounded question-answering in long egocentric videos.
\newblock In \emph{CVPR}, pages 12934--12943, 2024.

\bibitem[Fan(2019)]{fan2019egovqa}
Chenyou Fan.
\newblock Egovqa-an egocentric video question answering benchmark dataset.
\newblock In \emph{ICCV Workshop}, 2019.

\bibitem[Fu et~al.(2024)Fu, Dai, Luo, Li, Ren, Zhang, Wang, Zhou, Shen, Zhang, et~al.]{fu2024video}
Chaoyou Fu, Yuhan Dai, Yondong Luo, Lei Li, Shuhuai Ren, Renrui Zhang, Zihan Wang, Chenyu Zhou, Yunhang Shen, Mengdan Zhang, et~al.
\newblock Video-mme: The first-ever comprehensive evaluation benchmark of multi-modal llms in video analysis.
\newblock \emph{arXiv preprint arXiv:2405.21075}, 2024.

\bibitem[Grauman et~al.(2022)Grauman, Westbury, Byrne, Chavis, Furnari, Girdhar, Hamburger, Jiang, Liu, Liu, et~al.]{grauman2022ego4d}
Kristen Grauman, Andrew Westbury, Eugene Byrne, Zachary Chavis, Antonino Furnari, Rohit Girdhar, Jackson Hamburger, Hao Jiang, Miao Liu, Xingyu Liu, et~al.
\newblock Ego4d: Around the world in 3,000 hours of egocentric video.
\newblock In \emph{CVPR}, pages 18995--19012, 2022.

\bibitem[Guo et~al.(2021)Guo, Wang, and Wang]{guo2021context}
Dan Guo, Hui Wang, and Meng Wang.
\newblock Context-aware graph inference with knowledge distillation for visual dialog.
\newblock \emph{IEEE TPAMI}, 44\penalty0 (10):\penalty0 6056--6073, 2021.

\bibitem[Gurari et~al.(2018)Gurari, Li, Stangl, Guo, Lin, Grauman, Luo, and Bigham]{gurari2018vizwiz}
Danna Gurari, Qing Li, Abigale~J Stangl, Anhong Guo, Chi Lin, Kristen Grauman, Jiebo Luo, and Jeffrey~P Bigham.
\newblock Vizwiz grand challenge: Answering visual questions from blind people.
\newblock In \emph{CVPR}, pages 3608--3617, 2018.

\bibitem[He et~al.(2024)He, Ye, Zhang, Liu, and Tao]{he2024gomatching}
Haibin He, Maoyuan Ye, Jing Zhang, Juhua Liu, and Dacheng Tao.
\newblock Gomatching: A simple baseline for video text spotting via long and short term matching.
\newblock \emph{arXiv preprint arXiv:2401.07080}, 2024.

\bibitem[Hong et~al.(2024)Hong, Wang, Ding, Yu, Lv, Wang, Cheng, Huang, Ji, Xue, et~al.]{hong2024cogvlm2}
Wenyi Hong, Weihan Wang, Ming Ding, Wenmeng Yu, Qingsong Lv, Yan Wang, Yean Cheng, Shiyu Huang, Junhui Ji, Zhao Xue, et~al.
\newblock Cogvlm2: Visual language models for image and video understanding.
\newblock \emph{arXiv preprint arXiv:2408.16500}, 2024.

\bibitem[Jahagirdar et~al.(2023)Jahagirdar, Mathew, Karatzas, and Jawahar]{jahagirdar2023understanding}
Soumya Jahagirdar, Minesh Mathew, Dimosthenis Karatzas, and CV Jawahar.
\newblock Understanding video scenes through text: Insights from text-based video question answering.
\newblock In \emph{ICCV}, pages 4646--4650, 2023.

\bibitem[Jia et~al.(2022)Jia, Lei, Zhu, and Huang]{jia2022egotaskqa}
Baoxiong Jia, Ting Lei, Song-Chun Zhu, and Siyuan Huang.
\newblock Egotaskqa: Understanding human tasks in egocentric videos.
\newblock In \emph{NeurIPS}, pages 3343--3360, 2022.

\bibitem[Li et~al.(2024{\natexlab{a}})Li, Zhang, Zhang, Guo, Zhang, Li, Zhang, Liu, and Li]{li2024llava}
Bo Li, Kaichen Zhang, Hao Zhang, Dong Guo, Renrui Zhang, Feng Li, Yuanhan Zhang, Ziwei Liu, and Chunyuan Li.
\newblock Llava-next: Stronger llms supercharge multimodal capabilities in the wild, 2024{\natexlab{a}}.

\bibitem[Li et~al.(2022{\natexlab{a}})Li, Liu, Guo, Yin, Jiang, Du, Du, Zhu, Lai, Hu, et~al.]{li2022pp}
Chenxia Li, Weiwei Liu, Ruoyu Guo, Xiaoting Yin, Kaitao Jiang, Yongkun Du, Yuning Du, Lingfeng Zhu, Baohua Lai, Xiaoguang Hu, et~al.
\newblock Pp-ocrv3: More attempts for the improvement of ultra lightweight ocr system.
\newblock \emph{arXiv preprint arXiv:2206.03001}, 2022{\natexlab{a}}.

\bibitem[Li et~al.(2025)Li, Chen, Cai, Chen, Hong, Chen, Shen, and Gan]{li2025flexattention}
Junyan Li, Delin Chen, Tianle Cai, Peihao Chen, Yining Hong, Zhenfang Chen, Yikang Shen, and Chuang Gan.
\newblock Flexattention for efficient high-resolution vision-language models.
\newblock In \emph{ECCV}, 2025.

\bibitem[Li et~al.(2024{\natexlab{b}})Li, Wang, He, Li, Wang, Liu, Wang, Xu, Chen, Luo, et~al.]{li2024mvbench}
Kunchang Li, Yali Wang, Yinan He, Yizhuo Li, Yi Wang, Yi Liu, Zun Wang, Jilan Xu, Guo Chen, Ping Luo, et~al.
\newblock Mvbench: A comprehensive multi-modal video understanding benchmark.
\newblock In \emph{CVPR}, pages 22195--22206, 2024{\natexlab{b}}.

\bibitem[Li et~al.(2022{\natexlab{b}})Li, Wang, Xiao, Ji, and Chua]{li2022invariant}
Yicong Li, Xiang Wang, Junbin Xiao, Wei Ji, and Tat-Seng Chua.
\newblock Invariant grounding for video question answering.
\newblock In \emph{CVPR}, pages 2928--2937, 2022{\natexlab{b}}.

\bibitem[Li et~al.(2023)Li, Wang, Xiao, Ji, and Chua]{li2023transformer}
Yicong Li, Xiang Wang, Junbin Xiao, Wei Ji, and Tat-Seng Chua.
\newblock Transformer-empowered invariant grounding for video question answering.
\newblock \emph{IEEE TPAMI}, 2023.

\bibitem[Lin et~al.(2024)Lin, Yin, Ping, Molchanov, Shoeybi, and Han]{lin2024vila}
Ji Lin, Hongxu Yin, Wei Ping, Pavlo Molchanov, Mohammad Shoeybi, and Song Han.
\newblock Vila: On pre-training for visual language models.
\newblock In \emph{CVPR}, pages 26689--26699, 2024.

\bibitem[Lin et~al.(2022)Lin, Wang, Soldan, Wray, Yan, Xu, Gao, Tu, Zhao, Kong, et~al.]{lin2022egocentric}
Kevin~Qinghong Lin, Jinpeng Wang, Mattia Soldan, Michael Wray, Rui Yan, Eric~Z Xu, Difei Gao, Rong-Cheng Tu, Wenzhe Zhao, Weijie Kong, et~al.
\newblock Egocentric video-language pretraining.
\newblock \emph{NeurIPS}, 35:\penalty0 7575--7586, 2022.

\bibitem[Liu et~al.(2024{\natexlab{a}})Liu, Li, Li, Li, Zhang, Shen, and Lee]{liu2024llavanext}
Haotian Liu, Chunyuan Li, Yuheng Li, Bo Li, Yuanhan Zhang, Sheng Shen, and Yong~Jae Lee.
\newblock Llava-next: Improved reasoning, ocr, and world knowledge, 2024{\natexlab{a}}.

\bibitem[Liu et~al.(2024{\natexlab{b}})Liu, Li, Huang, Yang, Yu, Li, Yin, lin Liu, Jin, and Bai]{liu2024ocrbenchhiddenmysteryocr}
Yuliang Liu, Zhang Li, Mingxin Huang, Biao Yang, Wenwen Yu, Chunyuan Li, Xucheng Yin, Cheng lin Liu, Lianwen Jin, and Xiang Bai.
\newblock Ocrbench: On the hidden mystery of ocr in large multimodal models, 2024{\natexlab{b}}.

\bibitem[Luo et~al.(2024)Luo, Zhou, Zhang, Zheng, Sun, and Ji]{luo2024feast}
Gen Luo, Yiyi Zhou, Yuxin Zhang, Xiawu Zheng, Xiaoshuai Sun, and Rongrong Ji.
\newblock Feast your eyes: Mixture-of-resolution adaptation for multimodal large language models.
\newblock \emph{arXiv preprint arXiv:2403.03003}, 2024.

\bibitem[Maaz et~al.(2023)Maaz, Rasheed, Khan, and Khan]{maaz2023video}
Muhammad Maaz, Hanoona Rasheed, Salman Khan, and Fahad~Shahbaz Khan.
\newblock Video-chatgpt: Towards detailed video understanding via large vision and language models.
\newblock \emph{arXiv preprint arXiv:2306.05424}, 2023.

\bibitem[Mangalam et~al.(2023)Mangalam, Akshulakov, and Malik]{mangalam2023egoschema}
Karttikeya Mangalam, Raiymbek Akshulakov, and Jitendra Malik.
\newblock Egoschema: A diagnostic benchmark for very long-form video language understanding.
\newblock In \emph{NeurIPS}, pages 46212--46244, 2023.

\bibitem[Mathew et~al.(2021)Mathew, Karatzas, and Jawahar]{mathew2021docvqa}
Minesh Mathew, Dimosthenis Karatzas, and CV Jawahar.
\newblock Docvqa: A dataset for vqa on document images.
\newblock In \emph{WACV}, pages 2200--2209, 2021.

\bibitem[Mathew et~al.(2022)Mathew, Bagal, Tito, Karatzas, Valveny, and Jawahar]{mathew2022infographicvqa}
Minesh Mathew, Viraj Bagal, Rub{\`e}n Tito, Dimosthenis Karatzas, Ernest Valveny, and CV Jawahar.
\newblock Infographicvqa.
\newblock In \emph{WACV}, pages 1697--1706, 2022.

\bibitem[Mishra et~al.(2019)Mishra, Shekhar, Singh, and Chakraborty]{mishra2019ocr}
Anand Mishra, Shashank Shekhar, Ajeet~Kumar Singh, and Anirban Chakraborty.
\newblock Ocr-vqa: Visual question answering by reading text in images.
\newblock In \emph{ICDAR}, pages 947--952, 2019.

\bibitem[OpenAI(2024{\natexlab{a}})]{achiam2023gpt}
OpenAI.
\newblock Gpt-4o system card.
\newblock 2024{\natexlab{a}}.

\bibitem[OpenAI(2024{\natexlab{b}})]{gpt4omini2024}
OpenAI.
\newblock Gpt-4o mini: advancing cost-efficient intelligence.
\newblock 2024{\natexlab{b}}.

\bibitem[Pramanick et~al.(2023)Pramanick, Song, Nag, Lin, Shah, Shou, Chellappa, and Zhang]{pramanick2023egovlpv2}
Shraman Pramanick, Yale Song, Sayan Nag, Kevin~Qinghong Lin, Hardik Shah, Mike~Zheng Shou, Rama Chellappa, and Pengchuan Zhang.
\newblock Egovlpv2: Egocentric video-language pre-training with fusion in the backbone.
\newblock In \emph{ICCV}, pages 5285--5297, 2023.

\bibitem[Radford et~al.(2021)Radford, Kim, Hallacy, Ramesh, Goh, Agarwal, Sastry, Askell, Mishkin, Clark, et~al.]{radford2021learning}
Alec Radford, Jong~Wook Kim, Chris Hallacy, Aditya Ramesh, Gabriel Goh, Sandhini Agarwal, Girish Sastry, Amanda Askell, Pamela Mishkin, Jack Clark, et~al.
\newblock Learning transferable visual models from natural language supervision.
\newblock In \emph{ICML}, pages 8748--8763, 2021.

\bibitem[Reid et~al.(2024)Reid, Savinov, Teplyashin, Lepikhin, Lillicrap, Alayrac, Soricut, Lazaridou, Firat, Schrittwieser, et~al.]{reid2024gemini}
Machel Reid, Nikolay Savinov, Denis Teplyashin, Dmitry Lepikhin, Timothy Lillicrap, Jean-baptiste Alayrac, Radu Soricut, Angeliki Lazaridou, Orhan Firat, Julian Schrittwieser, et~al.
\newblock Gemini 1.5: Unlocking multimodal understanding across millions of tokens of context.
\newblock \emph{arXiv preprint arXiv:2403.05530}, 2024.

\bibitem[Schuhmann et~al.(2021)Schuhmann, Vencu, Beaumont, Kaczmarczyk, Mullis, Katta, Coombes, Jitsev, and Komatsuzaki]{schuhmann2021laion}
Christoph Schuhmann, Richard Vencu, Romain Beaumont, Robert Kaczmarczyk, Clayton Mullis, Aarush Katta, Theo Coombes, Jenia Jitsev, and Aran Komatsuzaki.
\newblock Laion-400m: Open dataset of clip-filtered 400 million image-text pairs.
\newblock \emph{arXiv preprint arXiv:2111.02114}, 2021.

\bibitem[Shang et~al.(2019)Shang, Di, Xiao, Cao, Yang, and Chua]{shang2019annotating}
Xindi Shang, Donglin Di, Junbin Xiao, Yu Cao, Xun Yang, and Tat-Seng Chua.
\newblock Annotating objects and relations in user-generated videos.
\newblock In \emph{Proceedings of the 2019 on International Conference on Multimedia Retrieval}, pages 279--287, 2019.

\bibitem[Shazeer(2020)]{shazeer2020glu}
Noam Shazeer.
\newblock Glu variants improve transformer.
\newblock \emph{arXiv preprint arXiv:2002.05202}, 2020.

\bibitem[Singh et~al.(2019)Singh, Natarajan, Shah, Jiang, Chen, Batra, Parikh, and Rohrbach]{singh2019towards}
Amanpreet Singh, Vivek Natarajan, Meet Shah, Yu Jiang, Xinlei Chen, Dhruv Batra, Devi Parikh, and Marcus Rohrbach.
\newblock Towards vqa models that can read.
\newblock In \emph{CVPR}, pages 8317--8326, 2019.

\bibitem[Sun et~al.(2023)Sun, Fang, Wu, Wang, and Cao]{sun2023eva}
Quan Sun, Yuxin Fang, Ledell Wu, Xinlong Wang, and Yue Cao.
\newblock Eva-clip: Improved training techniques for clip at scale.
\newblock \emph{arXiv preprint arXiv:2303.15389}, 2023.

\bibitem[Team et~al.(2023)Team, Anil, Borgeaud, Alayrac, Yu, Soricut, Schalkwyk, Dai, Hauth, Millican, et~al.]{team2023gemini}
Gemini Team, Rohan Anil, Sebastian Borgeaud, Jean-Baptiste Alayrac, Jiahui Yu, Radu Soricut, Johan Schalkwyk, Andrew~M Dai, Anja Hauth, Katie Millican, et~al.
\newblock Gemini: a family of highly capable multimodal models.
\newblock \emph{arXiv preprint arXiv:2312.11805}, 2023.

\bibitem[Tom et~al.(2023)Tom, Mathew, Garcia-Bordils, Karatzas, and Jawahar]{tom2023reading}
George Tom, Minesh Mathew, Sergi Garcia-Bordils, Dimosthenis Karatzas, and CV Jawahar.
\newblock Reading between the lanes: Text videoqa on the road.
\newblock In \emph{ICDAR}, pages 137--154, 2023.

\bibitem[Touvron et~al.(2023)Touvron, Martin, Stone, Albert, Almahairi, Babaei, Bashlykov, Batra, Bhargava, Bhosale, et~al.]{touvron2023llama}
Hugo Touvron, Louis Martin, Kevin Stone, Peter Albert, Amjad Almahairi, Yasmine Babaei, Nikolay Bashlykov, Soumya Batra, Prajjwal Bhargava, Shruti Bhosale, et~al.
\newblock Llama 2: Open foundation and fine-tuned chat models.
\newblock \emph{arXiv preprint arXiv:2307.09288}, 2023.

\bibitem[Wang et~al.(2024)Wang, Bai, Tan, Wang, Fan, Bai, Chen, Liu, Wang, Ge, et~al.]{wang2024qwen2}
Peng Wang, Shuai Bai, Sinan Tan, Shijie Wang, Zhihao Fan, Jinze Bai, Keqin Chen, Xuejing Liu, Jialin Wang, Wenbin Ge, et~al.
\newblock Qwen2-vl: Enhancing vision-language model's perception of the world at any resolution.
\newblock \emph{arXiv preprint arXiv:2409.12191}, 2024.

\bibitem[Wang et~al.(2023)Wang, Lv, Yu, Hong, Qi, Wang, Ji, Yang, Zhao, Song, et~al.]{wang2023cogvlm}
Weihan Wang, Qingsong Lv, Wenmeng Yu, Wenyi Hong, Ji Qi, Yan Wang, Junhui Ji, Zhuoyi Yang, Lei Zhao, Xixuan Song, et~al.
\newblock Cogvlm: Visual expert for pretrained language models.
\newblock \emph{arXiv preprint arXiv:2311.03079}, 2023.

\bibitem[Wang et~al.(2020)Wang, Liu, Shen, Ng, Luo, Jin, Chan, Hengel, and Wang]{wang2020general}
Xinyu Wang, Yuliang Liu, Chunhua Shen, Chun~Chet Ng, Canjie Luo, Lianwen Jin, Chee~Seng Chan, Anton van~den Hengel, and Liangwei Wang.
\newblock On the general value of evidence, and bilingual scene-text visual question answering.
\newblock In \emph{CVPR}, pages 10126--10135, 2020.

\bibitem[Wong et~al.(2022)Wong, Chen, Wu, Lei, Mao, Gao, and Shou]{wong2022assistq}
Benita Wong, Joya Chen, You Wu, Stan~Weixian Lei, Dongxing Mao, Difei Gao, and Mike~Zheng Shou.
\newblock Assistq: Affordance-centric question-driven task completion for egocentric assistant.
\newblock In \emph{ECCV}, pages 485--501. Springer, 2022.

\bibitem[Xiao et~al.(2021)Xiao, Shang, Yao, and Chua]{xiao2021next}
Junbin Xiao, Xindi Shang, Angela Yao, and Tat-Seng Chua.
\newblock Next-qa: Next phase of question-answering to explaining temporal actions.
\newblock In \emph{CVPR}, pages 9777--9786, 2021.

\bibitem[Xiao et~al.(2024)Xiao, Huang, Qin, Li, Li, Zhu, Tao, Yu, Lin, Chua, and Yao]{xiao2024videoqa}
Junbin Xiao, Nanxin Huang, Hangyu Qin, Dongyang Li, Yicong Li, Fengbin Zhu, Zhulin Tao, Jianxing Yu, Liang Lin, Tat-Seng Chua, and Angela Yao.
\newblock Videoqa in the era of llms: An empirical study.
\newblock \emph{arXiv preprint arXiv:2408.04223}, 2024.

\bibitem[Yang et~al.(2021)Yang, Feng, Ji, Wang, and Chua]{yang2021deconfounded}
Xun Yang, Fuli Feng, Wei Ji, Meng Wang, and Tat-Seng Chua.
\newblock Deconfounded video moment retrieval with causal intervention.
\newblock In \emph{Proceedings of the 44th international ACM SIGIR conference on research and development in information retrieval}, pages 1--10, 2021.

\bibitem[Yang et~al.(2022)Yang, Wang, Dong, Dong, Wang, and Chua]{yang2022video}
Xun Yang, Shanshan Wang, Jian Dong, Jianfeng Dong, Meng Wang, and Tat-Seng Chua.
\newblock Video moment retrieval with cross-modal neural architecture search.
\newblock \emph{IEEE Transactions on Image Processing}, 31:\penalty0 1204--1216, 2022.

\bibitem[Yang et~al.(2024)Yang, Zeng, Guo, Wang, Dong, and Wang]{yang2024robust}
Xun Yang, Jianming Zeng, Dan Guo, Shanshan Wang, Jianfeng Dong, and Meng Wang.
\newblock Robust video question answering via contrastive cross-modality representation learning.
\newblock \emph{Science China Information Sciences}, 67\penalty0 (10):\penalty0 202104, 2024.

\bibitem[Yao et~al.(2024)Yao, Yu, Zhang, Wang, Cui, Zhu, Cai, Li, Zhao, He, et~al.]{yao2024minicpm}
Yuan Yao, Tianyu Yu, Ao Zhang, Chongyi Wang, Junbo Cui, Hongji Zhu, Tianchi Cai, Haoyu Li, Weilin Zhao, Zhihui He, et~al.
\newblock Minicpm-v: A gpt-4v level mllm on your phone.
\newblock \emph{arXiv preprint arXiv:2408.01800}, 2024.

\bibitem[Ye et~al.(2024)Ye, Zhang, Daxberger, Chen, Lin, Li, Zhang, You, Xu, Gan, Lu, and Yang]{ye2024mm}
Hanrong Ye, Haotian Zhang, Erik Daxberger, Lin Chen, Zongyu Lin, Yanghao Li, Bowen Zhang, Haoxuan You, Dan Xu, Zhe Gan, Jiasen Lu, and Yinfei Yang.
\newblock Mm-ego: Towards building egocentric multimodal llms.
\newblock \emph{arXiv preprint arXiv:2410.07177}, 2024.

\bibitem[Zhai et~al.(2023)Zhai, Mustafa, Kolesnikov, and Beyer]{zhai2023sigmoid}
Xiaohua Zhai, Basil Mustafa, Alexander Kolesnikov, and Lucas Beyer.
\newblock Sigmoid loss for language image pre-training.
\newblock In \emph{ICCV}, pages 11975--11986, 2023.

\bibitem[Zhang et~al.(2024{\natexlab{a}})Zhang, Liu, Liu, Song, Wang, and Nie]{zhangmulti}
Haoyu Zhang, Meng Liu, Zixin Liu, Xuemeng Song, Yaowei Wang, and Liqiang Nie.
\newblock Multi-factor adaptive vision selection for egocentric video question answering.
\newblock In \emph{ICML}, 2024{\natexlab{a}}.

\bibitem[Zhang et~al.(2024{\natexlab{b}})Zhang, Zhou, Chen, Gu, Chen, and Sun]{zhang2024llava}
Ruiyi Zhang, Yufan Zhou, Jian Chen, Jiuxiang Gu, Changyou Chen, and Tong Sun.
\newblock Llava-read: Enhancing reading ability of multimodal language models.
\newblock \emph{arXiv preprint arXiv:2407.19185}, 2024{\natexlab{b}}.

\bibitem[Zhang et~al.(2024{\natexlab{c}})Zhang, Li, Liu, Lee, Gui, Fu, Feng, Liu, and Li]{zhang2024llavanextvideo}
Yuanhan Zhang, Bo Li, haotian Liu, Yong~jae Lee, Liangke Gui, Di Fu, Jiashi Feng, Ziwei Liu, and Chunyuan Li.
\newblock Llava-next: A strong zero-shot video understanding model, 2024{\natexlab{c}}.

\bibitem[Zhang et~al.(2024{\natexlab{d}})Zhang, Zhang, Li, Wang, Hou, Zou, and Bian]{zhang2024diffusion}
Yuzhe Zhang, Jiawei Zhang, Hao Li, Zhouxia Wang, Luwei Hou, Dongqing Zou, and Liheng Bian.
\newblock Diffusion-based blind text image super-resolution.
\newblock In \emph{CVPR}, pages 25827--25836, 2024{\natexlab{d}}.

\bibitem[Zhang et~al.(2024{\natexlab{e}})Zhang, Wen, Fu, Wang, Zhang, Wang, and Jin]{zhang2024beyond}
Yi-Fan Zhang, Qingsong Wen, Chaoyou Fu, Xue Wang, Zhang Zhang, Liang Wang, and Rong Jin.
\newblock Beyond llava-hd: Diving into high-resolution large multimodal models.
\newblock \emph{arXiv preprint arXiv:2406.08487}, 2024{\natexlab{e}}.

\bibitem[Zhao et~al.(2022)Zhao, Li, Wang, Li, Zhou, Zhang, Xuyang, Yu, Yu, Li, et~al.]{zhao2022towards}
Minyi Zhao, Bingjia Li, Jie Wang, Wanqing Li, Wenjing Zhou, Lan Zhang, Shijie Xuyang, Zhihang Yu, Xinkun Yu, Guangze Li, et~al.
\newblock Towards video text visual question answering: Benchmark and baseline.
\newblock In \emph{NeurIPS}, pages 35549--35562, 2022.

\bibitem[Zhou et~al.(2023)Zhou, Guo, Li, Yang, and Wang]{zhou2023exploring}
Sheng Zhou, Dan Guo, Jia Li, Xun Yang, and Meng Wang.
\newblock Exploring sparse spatial relation in graph inference for text-based vqa.
\newblock \emph{IEEE TIP}, 32:\penalty0 5060--5074, 2023.

\bibitem[Zhou et~al.(2024)Zhou, Xiao, Yang, Song, Guo, Yao, Wang, and Chua]{zhou2024scene}
Sheng Zhou, Junbin Xiao, Xun Yang, Peipei Song, Dan Guo, Angela Yao, Meng Wang, and Tat-Seng Chua.
\newblock Scene-text grounding for text-based video question answering.
\newblock \emph{arXiv preprint arXiv:2409.14319}, 2024.

\end{thebibliography}
